\documentclass[final]{siamonline0516}

\usepackage[top=0.9in,bottom=1.04in,left=1in,right=1in]{geometry}

\usepackage[backend=bibtex,firstinits=true,maxnames=6,url=false,isbn=false,doi=false]{biblatex}
\bibliography{Reference}

\usepackage{amsmath,amssymb,color,multirow}

\usepackage{verbatim}
\usepackage{algorithm,algorithmic}
\usepackage{fancybox}
\usepackage{fancyhdr}
\numberwithin{equation}{section}

\DeclareMathOperator*{\argmax}{arg\,max}
\DeclareMathOperator*{\argsup}{arg\,sup}

\newcommand{\RR}{\mathbb{R}}
\newcommand{\PP}{\mathbb{P}}
\newcommand{\EE}{\mathbb{E}}

\newcommand{\XX}{\mathcal{X}}
\newcommand{\OO}{\mathcal{O}}
\newcommand{\TT}{\mathcal{T}}
\newcommand{\LL}{\mathcal{L}}
\newcommand{\FF}{\mathcal{F}}

\newcommand{\ZZ}{\mathcal{Z}}
\newcommand{\CC}{\mathcal{C}}
\newcommand{\cK}{\mathcal{K}}
\newcommand{\cM}{\mathcal{M}}

\newcommand{\eps}{\epsilon}

\newcommand{\bd}[1]{\boldsymbol{#1}}

\newcommand{\mc}[1]{\mathcal{#1}}
\newcommand{\mk}[1]{\mathfrak{#1}}
\newcommand{\myM}{M}

\newcommand{\ud}{\,\mathrm{d}}

\newcommand{\todo}[1]{\textcolor{red}{(To Do: #1)}}
\newcommand{\red}[1]{{#1}}
\DeclareMathOperator\diag{diag}
\DeclareMathOperator\arginf{arg\ inf}

\newtheorem{theo}{Theorem}[section]

\newtheorem{cor}[theo]{Corollary}
\newtheorem{pro}[theo]{Proposition}
\newtheorem{remark}[theo]{Remark}

\title{Sequential Design for Ranking Response Surfaces}
\author{Ruimeng Hu and Mike Ludkovski \footnotemark[1]}

\begin{document}
\newcommand{\slugmaster}{}%

\maketitle
\footnotetext[1]{Department of  Statistics and Applied Probability
  University of California, Santa Barbara 93106-3110 \email{hu@pstat.ucsb.edu,ludkovski@pstat.ucsb.edu}. Work partially supported by NSF ATD-1222262.}

\markboth{Ruimeng Hu and Michael Ludkovski}{Sequential Design for Ranking Response Surfaces}

\begin{abstract}
Motivated by the problem of estimating optimal feedback policy maps in stochastic control applications, we
 propose and analyze sequential design methods for ranking several response surfaces. Namely, given $L \ge 2$ response surfaces over a continuous input space $\XX$, the aim is to efficiently find the index of the minimal response across the entire $\XX$. The response surfaces are not known and have to be noisily sampled one-at-a-time, requiring joint experimental design both in space and response-index dimensions. To generate sequential design heuristics we investigate Bayesian stepwise uncertainty reduction approaches, as well as sampling based on posterior classification complexity. We also
make  connections between our continuous-input formulation and the discrete framework of pure regret in multi-armed bandits. To model the response surfaces we utilize kriging metamodels. Several numerical examples using both synthetic data and an epidemics control problem are provided to illustrate our approach and the efficacy of respective adaptive designs.
\end{abstract}


\pagestyle{myheadings}
\thispagestyle{plain}

\section{Introduction}

A central step in stochastic control problems concerns estimating \emph{expected costs-to-go} that are used to approximate the optimal feedback control. In simulation approaches to this question, costs-to-go are sampled by generating trajectories of the stochastic system and then regressed against current system state. The resulting Q-values are finally ranked to find the action that minimizes expected costs.

\red{
When simulation is expensive, computational efficiency and experimental design become important. Sequential strategies rephrase learning the costs-to-go as another dynamic program, with actions corresponding to the sampling decisions.  In this article, we explore a Bayesian formulation of this sequential design problem. The ranking objective  imposes a novel loss function which mixes classification and regression criteria. Moreover, the presence of multiple stochastic samplers (one for each possible action) and a continuous input space necessitates development of targeted response surface methodologies. In particular, a major innovation is modeling in parallel the spatial correlation within each Q-value, while utilizing a  multi-armed bandit perspective for picking which sampler to call next.}

\red{To obtain a tractable approximation of the Q-values, we advocate the use of Gaussian process metamodels, viewing the latent response surfaces as realizations of a Gaussian random field. Consequently, the ranking criterion is formulated in terms of the posterior uncertainty about each Q-value. Thus, we connect metamodel uncertainty to the sampling decisions, akin to the discrete-state frameworks of ranking-and-selection and multi-armed bandits. Our work brings forth a new link between emulation of stochastic simulators and stochastic control, offering a new class of approximate dynamic programming algorithms.}

\subsection{Abstract Ranking Problem}

Let $\mu_\ell : \XX \to \RR$, $\ell \in \mk{L} \equiv \{1,2, \ldots, L\}$ be $L$ smooth functions over a subset $\XX$ of $\RR^d$. We are interested in the problem of learning the resulting \emph{ranking} of $\mu_\ell$ over the input space $\XX$, namely finding the classifier
\begin{equation}\label{def_cal}
\mathcal{C}(x) := \arg\min_\ell \left\{\mu_\ell(x)\right\} \in \mk{L}.
\end{equation}

The functions $\mu_\ell$ are a priori unknown but can be noisily sampled. That is for any $x \in \XX, \ell \in \mk{L}$ we have access to a simulator $Y_\ell(x)$ which generates estimates of $\mu_\ell(x)$:
\begin{equation}\label{def_Y}
Y_\ell(x) = \mu_\ell(x) + \eps_\ell(x), \ \ell \in \mk{L}
\end{equation}
where $\eps_\ell$ are independent, mean zero random variables with variance $\sigma_\ell^2(x)$. 
Intuitively speaking, we have $L$ smooth hyper-surfaces on $\XX$ that can be sampled via Monte Carlo. In the dynamic programming context, $x$ is the system state, $\ell$ indexes the various actions available to the controller, $\mu_\ell(\cdot)$ represents the expected costs-to-go and $\eps_\ell(\cdot)$ captures the simulation noise arising from pathwise simulation of the underlying stochastic system and corresponding costs.


Our goal is to identify the minimal surface \emph{globally} over the entire input space. More precisely, we seek to assign at each $x\in \XX$ a label $\hat{\CC}(x)$, while optimizing the loss metric
\begin{equation}\label{eq:loss}
\LL(\hat{\CC}, \CC) := \int_\XX \left\{ \mu_{\hat{\CC}(x)}(x) -\mu_{\CC(x)}(x) \right\} \; F(\ud x),
\end{equation}
\red{where $F(\cdot)$ is a specified weight function on $\XX$ determining the relative importance of ranking different regions. Thus, the loss is zero if the ranking is correct $\hat{\CC}(x) = \CC(x)$, and otherwise is proportional to the (positive) difference between the selected response and the true minimum $\mu_{\hat{\CC}}-\mu_{\CC}$.  The above criterion aims to identify the optimal action $\ell^*(x) \equiv \CC(x)$ to take in state $x$; if the wrong action $\hat{\CC}(x)$ is chosen instead, then \eqref{eq:loss} captures the resulting integrated loss to the controller assuming a probability distribution $F(\cdot)$ of potential states $x$.}

The loss function in \eqref{eq:loss} blends regression and classification objectives. In regression, one seeks to estimate the response marginally with the loss function tied to a single surface $\mu_\ell(\cdot)$. Instead, \eqref{eq:loss} is only about correctly identifying the index of the minimal response. As a result, small estimation errors are tolerated as long as the minimal response does not change, leading to a thresholding behavior in the loss function. In classification the loss function is discrete (typically with fixed mis-classification penalties), whereas \eqref{eq:loss} takes losses \emph{proportional} to the mis-classification distance $\mu_{\hat{\CC}(x)}(x)-\mu_{{\CC}(x)}(x)$.
 A further key distinction is that in classification the sampling space is just $\XX$ (returning a noisy label $C(x) \in \mk{L}$), whereas in our context a sampling query consists of the \emph{location-index} pair $(x,\ell) \in \XX \times \mk{L}$, sampling one response at a time. The question of which surface to sample requires separate analysis over $\mk{L}$.

 \red{We analyze  the design problem of constructing efficient sampling strategies that can well-estimate $\CC(\cdot)$ while optimizing the number of Monte Carlo samples needed. Because $\mu_\ell(\cdot)$'s are unknown, we frame \eqref{eq:loss} as a Bayesian sequential learning problem of adaptively growing a design $\ZZ$ that quickly learns $\CC(x)$. Classical static, i.e.~response-independent, designs are inadequate for ranking since the whole essence of optimizing computational efforts is predicated on \emph{learning} the structure of the unknown $\mu_\ell$'s. Intuitively, learning manifests itself in focusing the sampling through discriminating both in the input space $\XX$ (focus on regions where identifying $\CC(x)$ is difficult) and in the sampling indices $\mk{L}$ (focus on the surfaces where $\mu_\ell$ is likely to be the smallest response).}

Due to the joint design space $\XX \times \mathcal{L}$, our problem allows for a dual interpretation. Fixing $\ell$, \eqref{def_cal} is about reconstructing an unknown response surface $x \mapsto \mu_\ell(x)$ through noisy samples. Aggregating the different response surfaces, sequential design over $\XX$ reduces to identifying the partitions of $\XX = \cup_{i=1}^L \mathcal{C}_i$ into the sets
\begin{align}
\mathcal{C}_i := \{ x : \mathcal{C}(x) = i \} = \{ x : \mu_{\CC(x)}(x) = \min_\ell \mu_\ell(x) = \mu_i(x) \}, \quad i =1,\ldots,L.
\end{align}
Because in interiors of the partitions $\CC_i$ the ranking $\CC(x)$ is easier to identify, the main problem is to find the partition boundaries $\partial \mathcal{C}_i$. \red{ As a result, \eqref{def_cal} is related to contour-finding, for which sequential design was studied in \cite{GL13,Picheny10,RanjanBingham08}. Standard contour-finding attempts to identify the level set $\{ \mu_1(x) = a \}$ of the response surface, which corresponds to $L=2$ with known $\mu_2(x) = a$ in \eqref{def_cal}. Hence, the analysis herein can be viewed as a multi-variate extension of contour finding. In turn, contour-finding generalizes the classical  objective of minimizing a noisy response, connecting to the expected improvement/information gain trade-off in simulation optimization. In particular, we re-purpose the active learning rules of \cite{cohn:1996,Mackay92}.}

\red{
Conversely, fixing $x$, the aim of determining the smallest response $\arg\min_\ell \mu_\ell(x)$ corresponds to the setting of multi-armed bandits (MAB). The bandit has $L$ arms and corresponding payoffs $\mu_\ell(x), \ell \in \mathcal{L}$, with the decision-theoretic objective \eqref{def_cal} known as the pure exploration problem  \cite{BubeckMunos11,BubeckMunos11X}. Decision policies for which arm to pull are usually expressed in terms of posterior mean and confidence about the respective payoff; this point of view motivates our use of Gap-Upper Confidence Bound (UCB) design strategies \cite{Auer02,SRKRKASE:12}. Compared to  this  literature, \eqref{eq:loss} contains two key differences. First, the loss function is a weighted pure-regret criterion which to our knowledge has never been used in MAB context. Second, instead of a single bandit with independent arms, we treat the radical extension to a continuum of bandits indexed by $x \in \XX$. Recently, \cite{GrunewalderAudibert10,GabillonBubeck11}  considered multiple bandits which can be viewed as \eqref{def_cal} with a discrete, non-metrized $\XX$. We generalize their setting to a continuous $\XX$, with a spatial correlation structure of the arms. }


\subsection{Summary of Approach}

To handle continuous state spaces $x \in \XX$ that appear in stochastic control, we adopt the framework of  kriging or Gaussian process (GP) regression for modeling the Q-values. In both contexts of Design of Experiments (DoE) and continuous MAB's, kriging models have emerged as perhaps the most popular framework \cite{WilliamsRasmussenBook}. In particular, kriging has been used extensively for sequential regression designs
as it allows an intuitive approach to borrow information spatially across samples to build global estimates of the entire response surface $\mu_\ell$. Two further advantages are the analytic structure of Gaussian processes that allows for analytic evaluation of many Expected Improvement criteria, and the ability to naturally transition between modeling of deterministic (noise-free) experiments where data needs to be interpolated, and stochastic simulators where  data smoothing is additionally required.

More generally, we suggest a Bayesian perspective to global ranking, viewing the response surfaces as realizations of a random variable taking values in a given function space. This offers a tractable quantification of posterior metamodel uncertainty and related sequential metrics for determining the minimum surface. Thus, we stress that kriging is not essential to implementation of our algorithms; for example competitive alternatives are available among tree-based models, such as dynamic trees \cite{GTP-trees11} and Bayesian trees \cite{chip:geor:mccu:2010}. Moreover, while classical kriging may not be flexible enough for some challenging problems, there are now several well-developed generalizations, including treed GPs \cite{tgpPackage}, local GPs \cite{gramacy:apley:2013}, and particle-based GPs \cite{GramacyPolson11}, all offering off-the-shelf use through public \texttt{R} packages. 

Following the Efficient Global Optimization approach \cite{JonesSchonlauWelch98}, we define expected improvement scores that blend together the local complexity of the ranking problem and the posterior variance of our estimates. In particular, we rely on the expected \emph{reduction} in posterior variance and borrow from the Stepwise Uncertainty Reduction criteria based on GP regression from \cite{PichenyGinsbourger13,ChevalierPicheny13}. We also investigate UCB-type heuristics \cite{Auer02} to trade-off exploration and exploitation objectives. Based on the above ideas, we obtain a number of fully sequential procedures that specifically target efficient learning of $\mathcal{C}(\cdot)$ over the entire design space $\XX$. Extensive numerical experiments are conducted to compare these proposals and identify the most promising solutions.

As explained, our algorithms are driven by the exploration-exploitation paradigm quantified in terms of (empirically estimated) local ranking complexity for $\CC(x)$ and confidence in the estimated $\hat{\CC}$.  To quantify the local ranking complexity, we use the \emph{gaps} $\Delta(x)$ \cite{GabillonBubeck11,CarpentierLazaric11,HoffmanDeFreitas13}. For any $x \in \XX$, denote by $\mu_{(1)}(x) < \mu_{(2)}(x) < \ldots < \mu_{(L)}(x)$ the ranked responses at $x$ and by $$\Delta(x) :=\mu_{(1)}(x) - \mu_{(2)}(x)$$ the gap between the best (smallest) and second-best response. $\Delta(x)$ measures the difficulty in ascertaining $\CC(x)$: for locations where $\mu_{(1)} - \mu_{(2)}$ is \emph{big}, we do not need high fidelity, since the respective minimal response surface is easy to identify; conversely for locations where $\mu_{(1)} - \mu_{(2)}$ is small we need more precision. Accordingly, we wish to preferentially sample where $\Delta(x)$ is small. This is operationalized by basing the experimental design decisions on the estimated gaps $\widehat{\Delta}(x)$.

\red{In terms of design over $\mk{L}$, exploration suggests to spend the budget on learning the responses offering the biggest information gain. Namely, substantial benefits are available by discriminating over the sampling indices $\ell$ through locally concentrating on the (two) most promising surfaces $\mu_{(1)}, \mu_{(2)}$.  This strategy is much more efficient than
the naive equal sampling of each $Y_\ell$}. In addition, since the noise level in $Y_\ell$ may vary with $\ell$ this must also be taken into account. Summarizing, our expected improvement metrics  blend empirical gaps $\widehat{\Delta}$ and empirical posterior uncertainty based on kriging variance $\delta_\ell(x)$, jointly discriminating across $\XX \times \mk{L}$.


Our contributions can be traced along three directions. First, we introduce and analyze a novel sequential design problem targeting the loss function \eqref{eq:loss}. This setting is motivated by dynamic programming algorithms where statistical response models have been widely applied since the late 1990s \cite{Egloff05,Longstaff}. Here we contribute to this literature by proposing a Bayesian sequential design framework that generates substantial computational savings. This aspect becomes especially crucial in complex models where simulation is expensive and forms the main computational bottleneck. Second, we generalize the existing literature on Bayesian optimization and contour-finding to the multi-surface setting, which necessitates constructing new EI measures that address joint design in space and index dimensions. We demonstrate that this allows for a double efficiency gain: both in $\XX$ and in $\mathcal{L}$. Third, we extend the multiple bandits problem of \cite{GabillonBubeck11}
to the case of a continuum of bandits, which requires building a full meta-model for the respective arm payoffs. Our construction offers an alternative to the recent work \cite{BubeckMunos11X} on $\mathcal{X}$-armed bandits  and opens new vistas regarding links between MAB and DoE.

%
%

Our approach also generalizes Gramacy and Ludkovski \cite{GL13}. The latter work proposed sequential design for the contour-finding case where the design is solely over the input space $\XX$. In that context \cite{GL13} introduced several EI heuristics and suggested the use of dynamic trees for the response modeling. The framework herein however requires a rather different approach, in particular we emphasize the bandit-inspired tools (such as UCB) that arise with simultaneous modeling of multiple response surfaces.

The rest of the paper is organized as follows. Section \ref{sec:model} describes the kriging response surface methodology that we employ, as well as some analytic formulas helpful in the context of ranking. Section \ref{sec_estEI} then develops the expected improvement heuristics for \eqref{def_cal}. Sections \ref{sec:toy} and \ref{sec:epi} illustrate the designed algorithms using synthetic data (where ground truth is known), and a case-study from epidemic management, respectively. Finally, Section \ref{sec:conclude} concludes.


\subsection{Connection to Stochastic Control}\label{sec:control}
\red{Consider the objective of minimizing total costs associated with a controlled state process $X$,}
\begin{align}
c(0; u_{0:T}) = \sum_{t=0}^T g(t, X_t, u_t)
\end{align}
on the horizon $\{0,1,\ldots,T\}$.  Above $g(t,x,u)$ encodes the stagewise running costs, $u_{0:T}$ is the control strategy taking values in the finite action
space $u_t \in \mk{L}$, and $X_t \equiv X^{u}_t$ is a stochastic discrete-time
Markov state process with state space $\mathcal{X} \subseteq
\mathbb{R}^d$. The dynamics of $X^u$ are of the form
$$X^{u}_{t+1} =
F(X_t,u_t, \xi_{t+1})$$  for some map $F: \mathcal{X} \times \mk{L} \times \RR \to \mathcal{X}$, where $\xi_{t+1}$ is a random independent  centered noise source.
The performance criterion optimizes \emph{expected} rewards, captured in the value function $V(0,x)$,
$$
V(t,x) := \inf_{u_{t:T} \in \mathcal{U}} \EE[ c(t; u_{t:T}) | X_t = x], \qquad t \in \{0,1,\ldots,T\}, x \in \XX,
$$
over all admissible closed-loop Markov strategies $u_{t:T} \in \mathcal{U}$.
Thus, at time $t$, the action $u_t \equiv u(t,X_t)$ is specified in feedback form as a function of current state $X_t$. The policy map $(t,x) \mapsto u^*(t,x)$ translates system states into actions and is related to the value function via the dynamic
programming equation (DPE):
\begin{align}\label{eq:dpe}
V(t,x) &= \min_{u \in \mk{L}} \left\{ g(t,x,u) + \EE_t \left[ V(t+1, X^u_{t+1}) \right](x) \right\} = \mu_{u^*}(x;t), \\
\text{with} & \quad  \qquad \mu_u(x; t) := g(t,x,u) + \EE_t[ V(t+1, X^u_{t+1})](x). \label{eq:cost-to-go}
\end{align}
The notation $\EE_t[\cdot ](x) \equiv \EE[ \cdot | X_t=x]$ is meant to emphasize averaging of the
stochastic future at $t+1$ based on time-$t$ information summarized by the system state $X_t=x$. The term $\mu_u(x; t)$ is the Q-value, providing the expected cost-to-go if one applies action $u \in \mathcal{L}$ at $X_t=x$.

\red{Solving the DPE is equivalent to computing the Q-values since by \eqref{eq:dpe}, $V(t,x) = \min_{\ell \in \mk{L}} \{ \mu_\ell(x; t) \}$. The ranking problem in \eqref{def_cal} is then known as the policy map $x \mapsto u^*(t,x)$ that partitions the state space $\mathcal{X}$ into
$L$ {action sets} $\mathcal{C}_i(t)$. 
Given $u^*(s,\cdot)$ for all $s=t+1,\ldots, T$ and all $x \in \XX$ (initialized via $V(T,x) = g(T,x)$), we observe that
\begin{align}\label{eq:ls}
\mu_u(x; t) = g(t,x,u)+\EE_t \left[ \sum_{n={t+1}}^T g(n,X^{\widetilde{u}}_{n},\widetilde{u}_{n} ) \right](x),
\end{align}
where $(\widetilde{u}_t)$ is a strategy that uses action $u$ at $t$ and $u^*(s,X_s)$ thereafter, $s>t$.  Indeed, the sum in \eqref{eq:ls} is precisely the random variable for the pathwise costs-to-go $c(t,u_{t:T})$.
The loss \eqref{eq:loss} is then the difference between acting optimally as $u^*(t,X_t)$ at $t$, vis-a-vis taking action $\ell$ (and then acting optimally for the rest of the future, $\{t+1,\ldots,T\}$), weighted by the distribution $F(\ud x)$ of $X_t$.}

The formulation \eqref{eq:ls} allows pursuit of \emph{policy search} methods by tying the accuracy in \eqref{eq:cost-to-go} not to the immediate fidelity of (estimated) Q-values $\mu_u(\cdot; t)$, but to the quality of the policy map ${u}^*(t,x)$. Namely, one iteratively computes approximate policy maps $\hat{u}(s,\cdot)$ for $s=T-1, T-2, \ldots$, using \eqref{eq:ls} to construct $\hat{u}(t,\cdot)$ based on $\{ \hat{u}(s,\cdot) : s > t\}$. Note that the original objective of finding $V(0,x)$ requires solving $T$ ranking problems of the form \eqref{def_cal}.

This approach to dynamic programming is especially attractive when the action space $\mk{L}$ is very small. A canonical example are optimal stopping problems where $\mk{L} = \{\mathrm{stop}, \mathrm{continue}\}$, i.e.~$L=2$. For a single stopping decision, the immediate reward $\mu_2(x;t)$ is typically given, leading to the case of estimating a single Q-value $\mu_1(x;t)$, see \cite{GL13}.  Multiple stopping problems where both $\mu_1$ and $\mu_2$ need to be estimated arise in the pricing of swing options
\cite{MeinshausenHambly04}, valuation of real options \cite{AidLangrene12}, and
optimizing of entry-exit trading strategies \cite{ZervosJohnson12}. The case $L>2$ was considered for valuation of
energy assets, especially gas storage \cite{Secomandi11}, that lead
to optimal switching problems. For example, storage decisions are usually modeled in terms of the triple alternative $L=3$ of $\{inject, do-nothing, withdraw\}$.  Small action spaces also arise in many
engineering settings, such as target tracking \cite{AndersonMilutinovic11,HLQ15}, and sensor management \cite{VeeravalliFuemmeler08}. 

\section{Statistical Model}\label{sec:model}

\subsection{Sequential Design}\label{sec:seq-design}
Fix a configuration $\{\mu_\ell, \ell=1,\ldots,L\}$ and corresponding classifier $\CC(\cdot)$.
A design of size $K$ is a collection $\ZZ^{(K)} := (x,\ell)^{1:K}$, $x\in \XX, \ell \in \mk{L}$, with superscripts denoting vectors. Fixing $\ZZ^{(K)}$, and conditioning on the corresponding samples $Y^{1:K} \equiv (Y_{\ell^{k}}(x^{k}))_{k=1}^K$,
let $\hat{\CC}^{(K)} \equiv \hat{\CC}(Y^{1:K}, \ZZ^{(K)})$ be an estimate of $\CC$.  We aim to minimize the expected loss  $\LL( \hat{\CC}(\cdot, \ZZ^{(K)}), \CC)$ over all designs of size $K$, i.e.~
\begin{align}\label{global-design}
\inf_{\ZZ: | \ZZ| = K} \EE \left[ \LL(\hat{\CC}(Y^{1:K}, \ZZ), \CC) \right],
\end{align}
where the expectation is over the sampled responses $Y^{1:K}$. To tackle \eqref{global-design} we utilize sequential algorithms that iteratively augment the designs $\ZZ$ as $Y$-samples are collected. 
The interim designs $\ZZ^{(k)}$ are accordingly indexed by their size $k$, where $k=K_0, K_0+1, \ldots, K$. At each step, a new location $(x^{k+1}, \ell^{k+1})$ is added and the estimate $\hat{\CC}^{(k+1)}$ is recomputed based on the newly obtained information. The overall procedure is summarized by the following pseudo-code:
\begin{enumerate}
  \item Initialize $\ZZ^{(K_0)}$ and $\hat{\CC}^{(K_0)}$
  \item LOOP for $k=K_0,\ldots$
  \begin{enumerate}
     \item Select a new location $(x^{k+1},\ell^{k+1})$ and sample corresponding $y^{k+1} := Y_{\ell^{k+1}}(x^{k+1})$
     \item Augment the design $\ZZ^{(k+1)} = \ZZ^{(k)} \cup \{ (x^{k+1},\ell^{k+1})\}$
     \item Update the classifier $\hat{\CC}^{(k+1)}= \hat{\CC}( Y^{1:(k+1)}, \ZZ^{(k+1)})$ by assimilating the new observation
  \end{enumerate}
  \item END Loop
\end{enumerate}

The basic greedy sampling algorithm adds locations with the aim of minimizing the myopic expected estimation error. More precisely, at step $k$, given design $\ZZ^{(k)}$ (and corresponding $Y^{1:k}$), the next pair $\left(x^{k+1},\ell^{k+1}\right)$ is chosen by
\begin{align}\label{seq-design}
\arginf_{(x^{k+1},\ell^{k+1}) \in \XX \times \mk{L}} \EE \left[ \LL( \hat{\CC}( Y^{1:(k+1)}, \ZZ^{(k+1)}), \CC) \right],
\end{align}
where the expectation is over the next sample $Y_{\ell^{k+1}}(x^{k+1})$. This leads to a simpler one-step-ahead optimization compared to the $K$-dimensional  (and typically we are looking at $K \gg 100$) formulation in \eqref{global-design}. Unfortunately, the optimization in
 \eqref{seq-design} is still generally intractable because it requires
\begin{itemize}
\item re-computing the full loss function $\LL(\cdot, \CC)$ at each step;
\item finding the expected change in $\hat{\CC}$ given $Y_{\ell^{k+1}}(x^{k+1})$;
\item integrating over the (usually unknown) distribution of $Y_{\ell^{k+1}}(x^{k+1})$;
\item optimizing over the full $d+1$-dimensional design space $\XX \times \mk{L}$.
\end{itemize}

We accordingly propose efficient numerical approximations to \eqref{seq-design}, relying on the twin ideas of (i) sequential statistical modeling (i.e.~computing and updating $\hat{\CC}$ as $\ZZ$ grows), and (ii) stochastic optimization (i.e.~identifying promising new design sites $(x,\ell)$).

\subsection{Response Surface Modeling}\label{sec_reg}
A key aspect of sequential design is adaptive assessment of approximation quality in order to maximize information gain from new samples. Consequently, measuring predictive uncertainty is central to picking $(x^{k+1}, \ell^{k+1})$. For that purpose, we use a Bayesian paradigm, treating $\mu_\ell$ as random objects. Hence, we work with a function space $\mathcal{M}$ and assume that $\mu_\ell \in \mathcal{M}$ with some prior distribution $\FF_0$.
Thus, for each $x$, $\mu_\ell(x)$ is a random variable whose posterior distribution is updated based on the collected information from samples $(x, \ell, y_\ell(x))$.  Given the information generated by the $k^{th}$-step design $\ZZ^{(k)}$, $\FF_k = \sigma\left\{Y_\ell(x): (x,\ell) \in \ZZ^{(k)}\right\}$, we define the posterior $\myM^{(k)}_\ell(x) \sim \mu_\ell(x) | \FF_k$. The random variable $\myM^{(k)}_\ell(x)$ is the belief about $\mu_\ell(x)$ conditional on $\FF_k$; its first two moments are referred to as the kriging mean and variance respectively,
\begin{align}
  \widehat{\mu}^{(k)}_\ell(x)  & := \EE[ \mu_\ell(x) | \FF_k],\\
\delta^{(k)}_\ell(x)^2 & :=\EE[ (\mu_\ell(x)-\widehat{\mu}^{(k)}_\ell(x))^2 | \FF_k].
\end{align}
We will use $\widehat{\mu}(x)$ as a point estimate of $\mu_\ell(x)$, and $\delta_\ell(x)$ as a basic measure of respective uncertainty. The overall global map $x \mapsto \myM^{(k)}_\ell(x)$ is called the $\ell^{th}$ kriging surface. Note that while there is a spatial correlation structure over $\XX$, we assume that observations are independent across $\mk{L}$ (so sample noise $\epsilon_\ell \perp\!\!\perp \mu_\ell$), so that the posteriors $\myM^{(k)}_\ell(x)$, $\ell=1,2,\ldots$ are independent.


The order statistics $\widehat{\mu}_{(1)}(x) \le \widehat{\mu}_{(2)}(x) \le \ldots$ describe the sorted posterior means at a fixed $x$. A natural definition is to announce the minimum estimated surface
\begin{equation}\label{def_estcal}
\hat{\CC}(x) := \arg\min_\ell\left\{\widehat{\mu}_\ell(x)\right\},
\end{equation}
i.e.~the estimated classifier $\hat{\CC}$ corresponds to the smallest posterior mean, so that $\widehat{\mu}_{\hat{\CC}(x)}(x) = \widehat{\mu}_{(1)}(x)$. On the other hand, the uncertainty about $\CC(x)$ can be summarized through the expected minimum of the posteriors $\myM_1, \myM_2, \ldots, \myM_L$,
\begin{align}\label{eq:min-gaussian}
m^{(k)}(x):=\EE[\myM^{(k)}_{(1)}]=\EE[ \min(\mu_1(x),\ldots,\mu_L(x)) | \FF_k].
\end{align}
Observe that $\EE[ \min_\ell \mu_\ell(x) | \FF_k] = m^{(k)}(x) \le \widehat{\mu}^{(k)}_{(1)} = \min_\ell \EE[ \mu_\ell(x) |\FF_k]$, and we accordingly define the M-gap (``M'' for minimum)
\begin{align}
\label{M-Gap} \cM(x) := \widehat\mu_{(1)}(x) -m(x) \ge 0.
\end{align}
 The M-gap measures the difference between expectation of the minimum and the minimum expected response, which precisely corresponds to the Bayesian expected loss at $x$ in \eqref{eq:loss}. This fact offers an empirical analogue $\mc{EL} (\hat{\CC})$ of the original loss function $\LL(\hat{\CC}, \CC)$ in \eqref{eq:loss},
%
%
\begin{align}\label{eq:emploss}
\mc{EL}(\hat\CC) := \int_{\XX} \cM(x)  F(\ud x).
\end{align}
The above formula translates the local accuracy of the kriging surface into a global measure of fidelity of the resulting classifier $\hat{\CC}$ and will be the main performance measure for our algorithms.

\subsection{Kriging}
The response surfaces are assumed to be smooth in $\XX$. As a result, information about $\mu_\ell(x')$ is also revealing about $\mu_\ell(x)$ for $x\neq x'$, coupling observations at different sites. To enforce such conditions without a parametric representation, 
we view each $\mu_\ell$ as a sample from a Gaussian process (GP).
A GP is specified by its trend or mean function $t_\ell(x) = \EE[ \mu_\ell(x)]$ and a covariance structure $\cK_\ell : \XX^2 \to \mathbb{R}$, with $\cK_\ell(x,x') = \EE[ (\mu_\ell(x)-t_\ell(x))(\mu_\ell(x')-t_\ell(x'))]$.
By specifying the correlation behavior, the kernel $\cK$ encodes the smoothness of the response surface. 

Fix the response surface index $\ell$ and let $\vec{y} = (y(x^{1}), \ldots, y(x^{n}))^T$ denote the observed samples at locations $\vec{x} = x^{1:n}$. These realizations are modeled as in \eqref{def_Y} with the response represented as
$$\mu_\ell(x) = t_\ell(x) + Z_\ell(x), $$
where $t_\ell(\cdot)$ is a fixed trend term and $Z_\ell(\cdot)$ is a realization of a Gaussian process. Given the samples $(x,y)^{1:n}$, the posterior of $\mu_\ell$ again forms a GP; in other words any collection $\myM^{(n)}_\ell(x'_1), \ldots, \myM^{(n)}_\ell(x'_k)$ is multivariate Gaussian with mean $\widehat{\mu}^{(n)}_\ell(x'_i)$, covariance $v^{(n)}_\ell(x'_i,x'_j)$, and variance $\delta^{(n)}_\ell(x'_i)^2$, specified by \cite[Sec.~2.7]{WilliamsRasmussenBook} (see also \cite{NelsonStaum10}): 
\begin{align}\label{eq:krig-mean}
  \widehat{\mu}^{(n)}_\ell(x'_i) &=  t_\ell(x'_i) + \vec{k}^{(n)}_\ell(x'_i)^T (\mathbf{K}_\ell + \mathbf{\Sigma}_\ell^{(n)})^{-1} (\vec{y} - \vec{t}_\ell^{(n)} ) \\ \label{eq:krig-cov}
  v^{(n)}_\ell(x'_i,x'_j) & = {\cK}_\ell(x'_i,x'_j) - \vec{k}_\ell^{(n)}(x'_i)^T (\mathbf{K}_\ell + \mathbf{\Sigma}_\ell^{(n)})^{-1} \vec{k}_\ell^{(n)}(x'_j) 
  \end{align}
with \begin{align*} \delta^{(n)}_\ell(x'_i)^2 & = v^{(n)}_\ell(x'_i,x'_i) \qquad \vec{t}^{(n)}_\ell = (t_\ell(x^{1}), \ldots, t_\ell(x^{n}))^T, \quad\text{and}\\
\vec{k}_\ell^{(n)}(x'_i) & = (\cK_\ell(x^{1},x'_i), \ldots, \cK_\ell(x^{n},x'_i) )^T,
\qquad \mathbf{\Sigma}_\ell^{(n)} := \diag( \sigma^2_\ell(x^{1}), \ldots, \sigma^2_\ell(x^{n})),
\end{align*}
and $\mathbf{K}_\ell$ is the $n \times n$ positive definite matrix $(\mathbf{K}_\ell)_{i,j} := \cK_\ell(x^{i}, x^{j})$, $1 \le i,j \le n$.
%
By independence across $\ell$, the vector of posteriors $\bd{\myM(x)}$ at a fixed $x$ satisfies
$$\bd{\myM(x)} \sim \mc{N}(\bd{\widehat{\mu}}(x), \bd{\Delta}(x)) \quad\text{with} \;\; \bd{\widehat{\mu}}(x) = \left[\widehat\mu_1(x), \ldots, \widehat\mu_L(x)\right]^T\!, \;\; \bd{\Delta}(x) = \text{diag}\left(\delta_1^2(x), \ldots, \delta_L^2(x)\right).$$

\red{A common choice is the Matern-5/2 kernel
 \begin{align}\label{eq:matern}
 \cK(x,x'; s, \theta) = s^2 \bigl( 1+ (\sqrt{5} + 5/3) \| x-x' \|^2_\theta \bigr) \cdot e^{ - \sqrt{5} \| x - x'\|_\theta}, \qquad  \|x\|_\theta = \sqrt{ x \diag \vec\theta x^T}.
 \end{align}
The length-scale parameter vector $\vec\theta$ controls the smoothness of members of $\mathcal{M}_\cK$, the smaller the rougher. The variance scalar parameter $s^2$ determines the amplitude of fluctuations in the response.} 

A major advantage of kriging for sequential design are \emph{updating formulas} that allow to efficiently assimilate new data points into an existing fit. Namely, if a new sample $(x,y)^{k+1}$ is added to an existing design $x^{1:k}$, the mean and kriging variance at location $x$ are updated via
\begin{align}\label{eq:updated-mean}
 \widehat{\mu}^{(k+1)}(x) &=  \widehat{\mu}^{(k)}(x) + \lambda(x, x^{k+1}; x^{1:k}) (y^{k+1} - \widehat{\mu}^{(k)}(x^{k+1})); \\ \label{eq:updated-var}
  \delta^{(k+1)}(x)^2 & = \delta^{(k)}(x)^2 - \lambda(x, x^{k+1}; x^{1:k})^2 [\sigma^2(x^{(k+1)}) - \widehat{\mu}^{(k)}(x^{k+1}) ],
\end{align}
where $\lambda(x,x^{k+1}; x^{1:k})$ is a weight function specifying the influence of the new sample at $x^{k+1}$ on $x$ (conditioned on existing design locations $x^{1:k}$).
In particular, the local reduction in posterior standard deviation at $x^{k+1}$ is proportional to the current $\delta^{(k)}(x^{k+1})$ \cite{GinsbourgerEmery14}:
\begin{align}\label{eq:krig-update}
  \frac{\delta^{(k+1)}(x^{k+1})}{\delta^{(k)}(x^{k+1})} =  \frac{ \sigma(x^{k+1}) }{ \sqrt{ \sigma^2(x^{k+1})+ \delta^{(k)}(x^{k+1})^2} }.
\end{align}
Note that the updated posterior variance $\delta^{(k+1)}(x)^2$ is a deterministic function of $x^{k+1}$ which is independent of $y^{k+1}$.

For our examples below, we have used the \textbf{DiceKriging} \texttt{R} package \cite{kmPackage-R} to compute \eqref{eq:krig-mean}. The software takes as input the location-index pairs $(x,\ell)^{1:n}$, the corresponding samples $y_\ell(x)^{1:n}$ and the noise levels $\sigma^2_{\ell^n}(x^n)$, as well as the kernel family (Matern-5/2 \eqref{eq:matern} by default) and trend basis functions $t^i_\ell(x)$ and runs an EM MLE algorithm to estimate the  hyper-parameters $s, \theta$ describing the kriging kernel $\cK_\ell$. 

\subsection{Summary Statistics for Ranking}

Given a fitted kriging surface $\myM_\ell(\cdot)$ (for notational convenience in this section we omit the indexing by the design size $k$), the respective classifier $\hat\CC$ is obtained as in \eqref{def_estcal}. Note that $\hat{\CC}(x)$ is not necessarily the MAP (maximum a posteriori probability) estimator, since the ordering of the posterior probabilities and posterior means need not match for $L > 2$. Two further quantities are of importance for studying the accuracy of $\hat{\CC}$: gaps and posterior probabilities. First, the gaps quantify the differences between the posterior means, namely
\begin{align} 
\widehat{\Delta}_\ell(x) &:= |\widehat{\mu}_\ell(x) - \min_{j \neq \ell} \widehat{\mu}_j(x)|,\\ \label{eq:Delta}
\widehat{\Delta}(x) &:= |\widehat{\mu}_{(1)}(x) - \widehat{\mu}_{(2)}(x)|,
\end{align}
where $\widehat{\mu}_{(1)} \le \widehat{\mu}_{(2)} \le \ldots \le  \widehat{\mu}_{(L)} $ are the ordered posterior means. Note that under $L=2$,
we have $\widehat{\Delta}_1(\cdot) \equiv \widehat{\Delta}_2(\cdot) = \widehat{\Delta}(\cdot)$ due to symmetry. 
Second, define the posterior probabilities for the minimal rank
\begin{align}
p_\ell(x) & := \PP\left(\mu_\ell(x) = \mu_{(1)}(x) | \FF_k\right) = \PP( \myM_\ell(x) = \min_j \myM_j(x)) .\label{eq:probOfwrong} 
\end{align}
We refer to $p_{(1)}(x) \ge p_{(2)}(x)\ge \ldots \ge p_{(L)}(x)$ as the decreasing ordered values of the vector $\vec{p}(x) := \{p_\ell(x)\}_{\ell=1}^L$, so that the index of $p_{(1)}(x)$ is the MAP estimate of the minimal response surface.
The following proposition provides a semi-analytic recursive formula to evaluate $\vec{p}(x)$ in terms of the kriging means and variances $(\widehat{\mu}_\ell(x), \delta^2_\ell(x))$.
%
%
\begin{pro}[Azimi \emph{et al.} \cite{AzFeFe:11}]\label{prop1}
If $ \bd{\myM}(x) \sim \mc{N}(\bd{\widehat{\mu}}(x), \bd{\Delta}(x))$, then for any $\ell \in \mk{L}$,
\begin{equation}\label{eq:prop}
p_\ell(x) = \PP \left(\myM_\ell(x) = \min_j \myM_j(x) \right) = \prod\limits_{j=1}^{L-1} \Phi \left(-r_j^{(\ell)} \right),
\end{equation}
where $\Phi(\cdot)$ is standard normal cdf, and $\bd{r}^{(\ell)} = \left[r_1,r_2, \ldots, r_{L-1}\right]^T = (A(\ell)\bd{\Delta}(x)A(\ell)^T)^{-1/2}A(\ell)\bd{\widehat\mu}(x)$, with $A(\ell)$ a $(L-1)\times L$ matrix defined via
\begin{equation*}
A(\ell)_{i,j} = \left\{ \begin{aligned}
  {1}& \quad \text{if} { \quad j = \ell,} \\
  {-1}&\quad \text{if}{\quad 1 \leq i = j < \ell,} \text{ or }{\quad \ell < i+1 = j \leq L,} \\
  {0}& \quad\text{otherwise.}
\end{aligned} \right.
\end{equation*}
\end{pro}
\begin{cor}
For $L =2$, we have $p_1(x) = \PP(\myM_1(x) \leq \myM_2(x)) = \Phi\left(\frac{\widehat{\mu}_2(x) - \widehat{\mu}_1(x)}{\sqrt{\delta_1^2(x) + \delta_2^2(x)}}\right)$, and $p_2(x) = 1-p_1(x)$.
\end{cor}

The next proposition provides another semi-analytic formula to evaluate $m(x)$ defined in \eqref{eq:min-gaussian}.

\begin{pro}\label{prop2}
Suppose that $L=2$ and
let $\myM_\ell(x) \sim \mc{N}(\widehat\mu_\ell(x), \delta^2_\ell(x))$, $\ell=1,2$ be two independent Gaussians. Define
\begin{align*}
d_{12} := \sqrt{\delta_1^2(x) + \delta_2^2(x)},\qquad\text{and} \quad
a_{12} := (\widehat\mu_1(x) - \widehat\mu_2(x))/d_{12}.
\end{align*}
Then  the first two moments of $\myM_{(1)}(x) = \min(\myM_1(x),\myM_2 (x))$ are given by:
\begin{align} \label{eq:min-mean}
m(x) & \equiv \EE[\myM_{(1)}(x)] = \widehat\mu_1(x)\Phi(-a_{12})+\widehat\mu_2(x)\Phi(a_{12}) - d_{12}\phi(a_{12}),\\ \label{eq:min-var}
\EE \left[\myM_{(1)}(x)^2 \right]  &= (\widehat\mu_1^2(x)+\delta_1^2(x))\Phi(-a_{12}) + (\widehat\mu_2^2(x) + \delta_2^2(x))\Phi(a_{12}) \\
\notag & \qquad  - (\widehat\mu_1(x)+\widehat\mu_2(x))d_{12}\phi(a_{12}).
\end{align}
\end{pro}
Equation \eqref{eq:min-mean} provides a closed-form expression to evaluate $m(x) = \EE[ \myM_{(1)}(x)]$ for $L=2$. In the case $L>2$, one may evaluate $m(x)$ recursively using a Gaussian approximation. For instance, for $L = 3$, approximate $\widetilde{Y} := \myM_1(x) \wedge \myM_2(x)$ by a Gaussian random variable with mean/variance specified by \eqref{eq:min-mean}-\eqref{eq:min-var} respectively (i.e.~using $a_{12}$ and $d_{12}$) and then apply Proposition \ref{prop2} once more to $\myM_{(1)}(x) = \widetilde{Y} \wedge \myM_3(x)$.


\section{Expected Improvement}\label{sec_estEI}
The Bayesian approach to sequential design is based on greedily optimizing an acquisition function. The optimization is quantified through Expected Improvement (EI) scores that identify pairs $(x,\ell)$ that are most promising in terms of lowering the global empirical loss function $\mc{EL}$ according to \eqref{seq-design}. In our context the EI scores are based on the posterior distributions $\myM_\ell^{(k)}$ which summarize information learned so far about $\mu_\ell(x)$.



Our two main heuristics are dubbed Gap-UCB and Gap-SUR:
\begin{align}
  E^{Gap-UCB}_k(x, \ell) & := -\widehat\Delta_\ell(x) + \gamma_k\delta_\ell(x); \label{eq:EGapSd} \\
E^{Gap-SUR}_k(x,\ell) & := \EE[ \cM^{(k)}(x) - \cM^{(k+1)}(x) | x^{k+1} =x, \ell^{k+1} = \ell, \FF_k]. \label{eq:MGap}
\end{align}
The Gap-UCB score is motivated by the exploration-exploitation trade-off in MAB's and favors locations with small gaps in posterior means and high kriging variance. Indeed, the local empirical gap measure \cite{GabillonBubeck11} $\widehat{\Delta}_\ell(x)$ identifies the most promising arm, while the kriging variance $\delta_\ell^2(x)$ promotes exploration to reduce uncertainty about arm payoffs. The two are connected via the UCB (upper confidence bound \cite{SRKRKASE:12}) tuning parameter  $\gamma_k$  that balances exploration (regions with high $\delta_\ell(x)$) and exploitation (regions with small gap).
Another interpretation of Gap-UCB is to mimic a complexity-sampling scheme that selects design sites based on the complexity of the underlying ranking problem. Indeed, the gap $\Delta_\ell(x) := {\mu_{\ell}(x) - \min_{j \neq \ell}\mu_j(x)}$ measures the hardness of testing whether $\mu_\ell(x) = \min_i \mu_i(x)$; the smaller $\Delta_\ell(x)$ the tougher. At the same time, the kriging variance $\delta^2(x)$ can be related to information gain from sampling at $x$ (being akin to the standard error of a point estimator).

The Gap-SUR strategy is coming from the perspective of simulation optimization.
 Recall that we strive to lower the empirical loss $\mc{EL}$ in \eqref{eq:emploss} which is related to the M-gap in \eqref{eq:MGap}, $\mc{EL} = \int \cM(x) F(\ud x)$. Accordingly, the Gap-SUR criterion uses $\cM(x)$ to guide the adaptive design, by aiming to maximizing its \emph{expected local reduction} if we add $(x,\ell)$ to the design. Such Stepwise uncertainty reduction (SUR) strategies were introduced in \cite{Picheny12,ChevalierPicheny13}.
  The evaluation of \eqref{eq:MGap} requires computing the expected mean and variance of $M_{(1)}(x)$ and $M_\ell(x)$. The updating formula \eqref{eq:updated-mean} implies that (keeping $\cK$ fixed) $\EE[\widehat{\mu}^{k+1}_\ell(x) |x^{k+1} = x, \ell^{k+1}=\ell, \FF_k ]  = \widehat{\mu}^{k}_\ell(x)$, while \eqref{eq:krig-update} yields
 $\delta_\ell^{(k+1)}(x)$. The rest of the computation becomes straightforward in view of Proposition \ref{prop2}.
\red{\begin{remark}
Gap-SUR is also connected to the Active Learning Cohn (ALC) \cite{cohn:1996} approach to DoE. In ALC, minimization of posterior variance is achieved by greedily maximizing reduction in $\delta^2(x)$. In Gap-SUR, minimization of $\mc{EL}$ is achieved by maximizing reduction in $\cM(x)$. The ALC paradigm suggests an alternative to \eqref{eq:EGapSd}, namely $E^{Gap-ALC}_k(x, \ell)  = -\widehat\Delta_\ell(x) + \gamma_k [\delta^{(k)}_\ell(x) - \delta^{(k+1)}_\ell(x)]$, that blends expected decline in kriging variance with the estimated gap.
%
\end{remark}}

\red{\textbf{Asymptotic Behavior.}
The Gap-SUR method aims to drive the M-gaps to zero, which is equivalent to learning all the responses: $\cM(x) = 0 \Leftrightarrow \delta_\ell(x)= 0 \;\forall \ell$, see \eqref{eq:MGap}.  For GP models, vanishing posterior variance at $x$ corresponds to the design being dense in the neighborhood of $x$. Thus, asymptotically, the Gap-SUR heuristic will generate designs that are dense across $\XX \times \mk{L}$. Finally, previous results about consistency of GP models (see for example \cite{ChoiSchervish07}) can be invoked to establish that $\hat{C} \to C$.}

\red{On the other hand, proper selection of the UCB schedule $(\gamma_k)$ is crucial for the performance of Gap-UCB. If $\gamma_k \equiv 0$ then convergence is not guaranteed. Indeed, consider $x_1,x_2$ such that $\Delta(x_2) > \Delta(x_1)$,
but the estimated gaps based on interim $\ZZ^{(k)}$ satisfy $\widehat{\Delta}(x_1)> \widehat{\Delta}(x_2) > \Delta(x_2)$ due to estimation error at $x_1$. Then at stage $k$ the algorithm will prefer site $x_2$ over $x_1$ (since it has smaller gap $\widehat{\Delta}$) and will then possibly get trapped indefinitely, never realizing that the estimated ordering between $\Delta(x_1)$ and $\Delta(x_2)$ is wrong. Hence without UCB the algorithm is prone to get trapped at local minima of $\widehat{\Delta}$. At the same time, any increasing unbounded $\gamma_k \to +\infty$ guarantees that $\sup_x \delta_\ell^{(k)}(x) \to 0 \; \forall \ell$.  Toward this end, Srinivas et al.~\cite{SRKRKASE:12}  proved that in a cumulative regret setting  $\gamma_k = O( \sqrt{\log k})$ should grow logarithmically in sample size $k$. Further rules on how to choose $\gamma_k$ (for the case of a finite state space $\XX$) can be found in \cite{GabillonBubeck11}. Another alternative is a randomized version. For example, in $\eps$-greedy sampling, with probability $\eps$ at any step instead of using an EI metric, $(x,\ell)^{k+1}$ are selected uniformly in $\XX \times \mk{L}$. This ensures that the designs $\ZZ^{(k)}$ are dense in $\XX$ as $k \to \infty$ and is a feature that we resort to in our experiments. Still, fine-tuning the schedule of $k \mapsto \gamma_k$ is highly non-trivial in black-box settings. For this reason, usage of the Gap-UCB approach is sensitive to implementation choices and further guidance on selecting $(\gamma_k)$ is left for future research.
}

\subsection{Selecting the Next Sample Location}\label{sec_loc}
To grow the designs $\ZZ^{(k)}$ over $k=K_0,K_0+1,\ldots$ we use the EI scores via the greedy sampling strategy
\begin{align}\label{eq:best-1}
  (x,\ell)^{k+1} = \argsup_{(x, \ell) \in \XX \times \mk{L}} E_k(x,\ell).
\end{align}
Because the above introduces a whole new optimization sub-problem, in cases where this is computationally undesirable we instead replace $\arg\sup_{ x \in \XX }$ with $\arg\max_{x \in \TT }$ where $\TT$ is a finite \emph{candidate set}. Optimization over $\TT$ is then done by direct inspection. The justification for this procedure is that (i) we expect $E_k(x, \ell)$ to be smooth in $x$ and moreover relatively flat around $x^*$; (ii) $E_k(x, \ell)$ is already an approximation so that it is not required to optimize it precisely; (iii) performance of optimal design should be insensitive to small perturbations of the sampling locations. To construct such candidate sets $\TT$ in $\XX$,  we employ Latin hypercube sampling (LHS) \cite{McBeCo:79}.
LHS candidates ensure that new locations are representative, and well spaced out over $\XX$. See \cite[Sec 3.4]{gra:lee:2009} for some discussion on how $\TT$ should be designed. In addition, we refresh our \emph{candidate set} $\TT$ at each iteration, to enable ``jittering''. Algorithm \ref{algorithm} below presents the resulting method in pseudo-code.

\begin{algorithm}[ht]
\caption{Sequential Design for Global Ranking using Kriging \label{algorithm}}
\begin{algorithmic}[1]
\REQUIRE $K_0, K$ 
\STATE Generate initial design $\ZZ^{(K_0)} := (x, \ell)^{1:K_0}$ using LHS
\STATE Sample $y^{1:K_0}$, estimate the GP kernels $\cK_\ell$'s and initialize the response surface models $\myM_\ell$
\STATE Construct the classifier $\mathcal{C}^{(K_0)}(\cdot)$ using \eqref{def_estcal}
\STATE $k \leftarrow K_0$
\WHILE{$k < K$}
\STATE Generate a new candidate set $\TT^{(k)}$ of size $D$
\STATE Compute the expected improvement (EI) $E_{k}(x,\ell)$ for each $x \in \TT$, $\ell \in \mk{L}$
\STATE Pick a new location $\displaystyle(x,\ell)^{k+1} = \argmax_{(x,\ell) \in \TT^{(k)}\times \mk{L}} E_k(x,\ell)$ and sample the corresponding $y^{k+1}$
\STATE (Optional) Re-estimate the kriging kernel $\cK_{\ell^{k+1}}$
\STATE Update the response surface $\myM_{\ell^{k+1}}$ using \eqref{eq:updated-mean}-\eqref{eq:updated-var}
\STATE Update the classifier $\CC^{(k+1)}$ using \eqref{def_estcal}
\STATE Save the overall grid $\ZZ^{(k+1)} \leftarrow \ZZ^{(k)} \cup (x^{k+1}, \ell^{k+1})$
\STATE k $\leftarrow$ k+1
\ENDWHILE
\RETURN Estimated classifier $\CC^{(K)}(\cdot)$.
\end{algorithmic}
\end{algorithm}

\begin{remark}\label{sec:implement}
\red{In the context of a kriging model, the initial design $\ZZ^{(K_0)}$ is crucial to allow the algorithm to learn the  covariance structures of the responses. One common challenge is to avoid assuming that $\mu_\ell$'s are too flat by missing the shorter-scale fluctuations \cite{Picheny10}. Thus, $K_0$ must be large enough to reasonably estimate $\cK_\ell$; one recommendation is that $K_0$ should be about 20\% of the eventual design size $K$. In our implementation, the initialization is done via a space-filling LHS design (sampling equally across the $L$ surfaces).  
Another issue is the re-estimation of the kriging kernel $\cK_\ell$ in step 9 of Algorithm \ref{algorithm}. Re-training is computationally expensive and makes the GP framework not sequential. 
Since we expect the algorithm to converge as $k \to \infty$, we adopt the practical rule of running the full estimation procedure for $\cK$ according to the doubling method \cite{GanoRenaud06}, re-estimating $\cK_\ell$ for $k=2,4,8,\ldots$ a power of two, and keeping it frozen otherwise.}
\end{remark}

\subsubsection{Hierarchical and Concurrent Sampling}\label{sec:two-step}
Instead of sampling directly over the pairs $(x,\ell) \in \XX \times \mk{L}$, one can consider two-step procedures that first pick $x$ and then $\ell$ (or vice-versa). This strategy matches standard sequential designs over $\XX$. Indeed, one can then directly follow the active learning approach of \cite{Mackay92,cohn:1996} by first picking $x^{k+1}$ using the gap metrics, and then picking the index $\ell^{k+1}$ based on the kriging variance:
\begin{align}\label{eq:two-step} \left\{ \begin{aligned}
  x^{k+1} &= \arg\min_{x\in\XX} \widehat{\Delta}(x) | \FF_k, \qquad \text{cf.}~\eqref{eq:Delta} \\
  \ell^{k+1} &= \arg\max_{\ell \in \mk{L}} \delta^{(k)}_\ell(x^{k+1}).
\end{aligned} \right.
\end{align}
Conditional on picking $x^{k+1}$, the above choice selects surfaces with large kriging variance $\delta_\ell(x)$, attempting to equalize $\delta_\ell(x)$ across $\ell$. \red{Note that \eqref{eq:two-step} will focus on the most \emph{uncertain} response, not on the most promising one, which tends to hurt overall performance when $L \gg 2$.} Another choice is to pick $\ell^{k+1}$ to greedily maximize the information gain as in \eqref{eq:krig-update}. Such two-step EI heuristics allow to avoid having to specify the  schedule $\gamma_k$ of UCB criteria \eqref{eq:EGapSd}.

\red{A further variant is concurrent marginal modeling of each $\mu_\ell(\cdot)$. This is achieved by concurrent sampling:} after choosing a location $x^{k+1} \equiv x$, one augments the design with the $L$ respective pairs $\left(x,1\right), \left(x,2\right), \ldots \left(x, L\right)$. This approach ``parallelizes'' the learning of all response surfaces while still building an adaptive design over $\XX$.
The disadvantage of this strategy becomes clear in the extreme situation when the variance of $Y_1(x)$ is zero, $\sigma_1(x) \equiv 0$ while the noise of $Y_2(x)$ is large. In that case, after sampling a given location once for each response, $(x,1)$ and $(x,2)$, we would have $\delta_1(x) = 0$, $\delta_2(x) \gg 0$. Hence, another sample from $Y_1(x)$ would gain no information at all, while substantial information would still be gleaned from sampling $Y_2(x)$, making parallel sampling twice as costly as needed.

\section{Simulated Experiments}\label{sec:toy}

\subsection{Toy Example}\label{sec:1d}

In this section we consider a simple one-dimensional example with synthetic data  which allows a fully controlled setting.
Let $L = 2, \XX = [0, 1]$. The noisy responses $Y_1(x)$ and $Y_2(x)$ are specified by (cf.~the example in \cite[Sec~4.4]{kmPackage-R})
\begin{align*}
 Y_1(x) &= \mu_1(x) + \eps_1(x) \equiv \frac{5}{8}\left(\frac{\sin(10x)}{1+x} + 2x^3\cos(5x)+0.841\right) + \sigma_1(x)Z_1, \\
 Y_2(x) &= \mu_2(x) + \eps_2(x) \equiv 0.5 + \sigma_2(x)Z_2.
\end{align*}
Here $Z_\ell$ are independent standard Gaussian, and the noise strengths are fixed at $\sigma_1(x) \equiv 0.2$ and $\sigma_2(x) \equiv 0.1$, homoscedastic in $x$ but heterogenous in $\ell=1,2$. The weights $F(\ud x)=\ud x$ in the loss function are uniform on $\XX$.
%
%
The true ranking classifier $\CC(x)$ is given by
\begin{equation}
\CC(x) = \left\{ {\begin{array}{*{20}{l}}
  {2}& \quad \text{for } x \in [0,r_1] \cup [r_2,1] \\
  {1}& \quad \text{for } {r_1 < x < r_2, } \\
\end{array}} \right.
\end{equation}
where $r_1 \approx 0.3193, r_2 \approx 0.9279.$

\begin{figure}[ht]
  \centering
 \begin{tabular}{cc}
   \includegraphics[width=0.48\textwidth,height=2.25in]{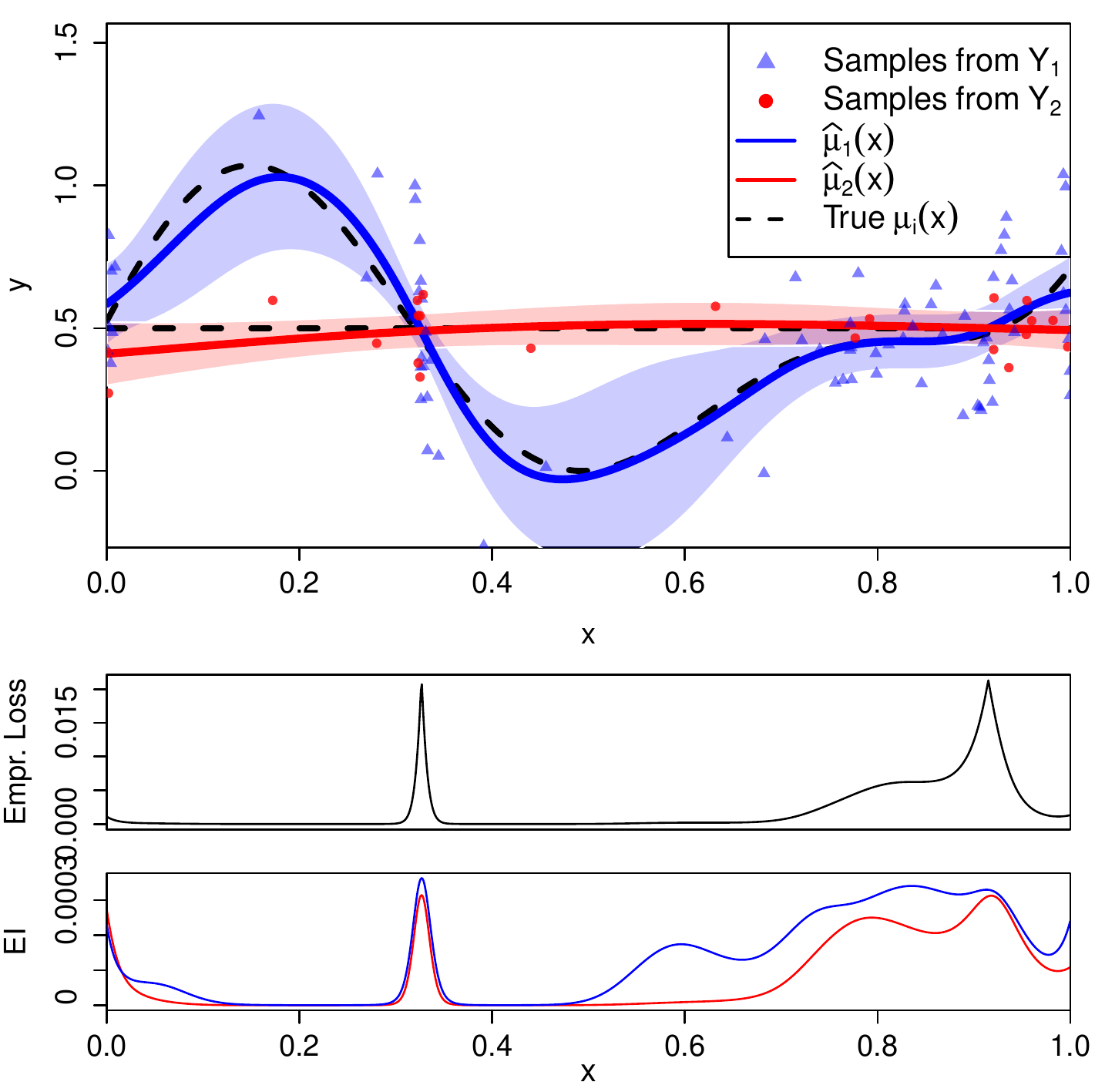} &
  \includegraphics[width=0.48\textwidth,height=2.25in]{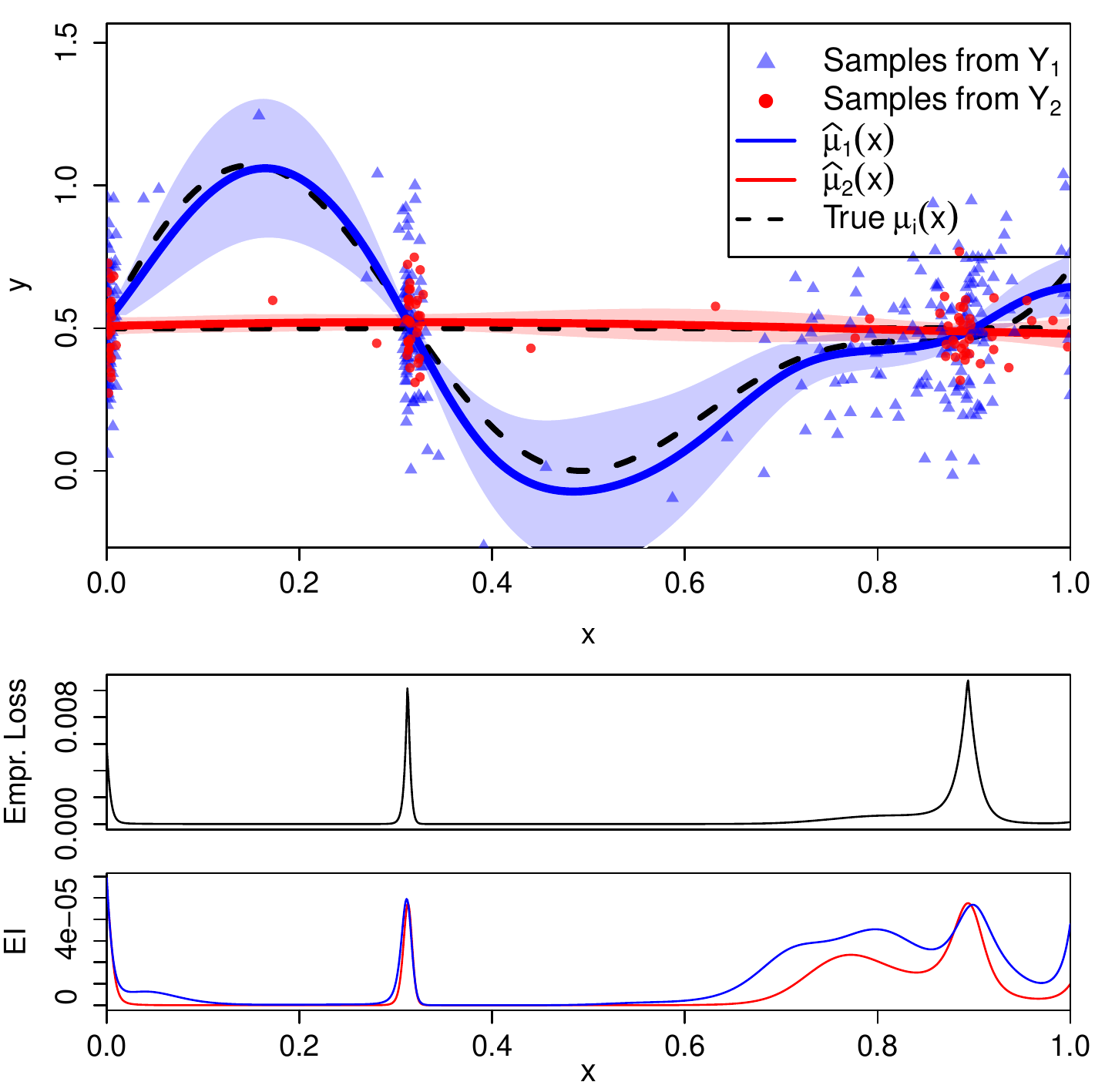} \\
  $K=100$ & $K=400$ \end{tabular}
  \caption{Response surface modeling with the Gap-SUR EI criterion of \eqref{eq:MGap}. We plot the true surfaces $\mu_\ell(x)$ (black dashed lines), the posterior means $\widehat{\mu}_\ell(x)$ (blue/red solid lines), the 90\% posterior credibility intervals (light blue/red areas) of $\myM_1(x)$ and $\myM_2(x)$, and the sampling locations $x^{1:K}$ for $Y_1(x)$ (blue triangles) and $Y_2(x)$ (red circles).  The middle panel shows the local loss $\cM(x)$, cf.~\eqref{M-Gap}, while the bottom panel shows the Gap-SUR EI metric $E_K(x,\ell)$ (blue: $\ell=1$, red: $\ell=2$).  \label{fig:conv-1d} }
\end{figure}

To focus on the performance of various acquisition functions, we fix the kriging kernels $\cK_\ell$ to be of the Matern-5/2 type \eqref{eq:matern} with hyperparameters $s_1=0.1, \theta_1 = 0.18$ for $\cK_1$ and $s_2=0.1, \theta_2=1$ for $\cK_2$. These hyper-parameters are close to those obtained by training a kriging model for $Y_\ell(x)$ given a dense design on $\XX$ and hence capture well the smoothness of the response surfaces above.
\red{We use a fixed trend $t_\ell(x) = 0.5$, and treat the given sampling noises $\sigma_\ell$ as known.}

%

To apply Algorithm \ref{algorithm} we then initialize with $K_0=10$ locations $(x,\ell)^{1:K_0}$ (five each from $Y_1(x)$ and $Y_2(x)$), drawn from a LHS design on $[0,1]$. Note that because the kriging kernels are assumed to be known, $K_0$ is taken to be very small. To grow the designs we employ the Gap-SUR EI criterion and optimize for the next $(x,\ell)^{k+1}$ using a fresh candidate set $\TT^{(k)}$ based on a LHS design of size $D=100$.
 %
%
%
Figure \ref{fig:conv-1d} illustrates the evolution of the posterior response surface models. The two panels show the estimated $\myM^{(K)}_\ell(x)$ at $K=100$ and $K=400$ (namely we plot the posterior means $\widehat{\mu}^{(K)}_\ell(x)$ and the corresponding 90\% CI $\widehat{\mu}^{(K)}_\ell(x) \pm  1.645\delta^{(K)}_\ell(x)$).  We observe that most of the samples are heavily concentrated around the two classification boundaries $r_1, r_2$, as well as the ``false'' boundary at $x=0$. As a result, the kriging variance $\delta^2_\ell(x)$ is much lower in those neighborhoods, generating the distinctive ``sausage'' shape for the posterior credibility intervals of $\myM_\ell(x)$. In contrast, in regions where the gap $\Delta(x)$ is large (e.g., around $x=0.5$), ranking the responses is easy so that almost no samples are taken and the kriging variance remains large. Also, because $\sigma_1(x) > \sigma_2(x)$, the credibility intervals of $\mu_2$ are tighter, $\delta_1(x) > \delta_2(x)$, and more than 70\% of the samples are from the first response $Y_1$. Indeed, we find $D_1(k) \simeq 3 D_2(k)$ where
\begin{align*}
  D_i(K) := | \{ 1 \le k \le K : \ell^k = i \}|
\end{align*}
is the number of samples in the design $\ZZ^{(K)}$ from the $i$-th surface. The above observations confirm the double efficiency from making the EI scores depend on both  the $\XX$ and $\mk{L}$ dimensions.

\begin{figure}[ht]
  \centering
 \includegraphics[width=0.48\textwidth,height=2.25in]{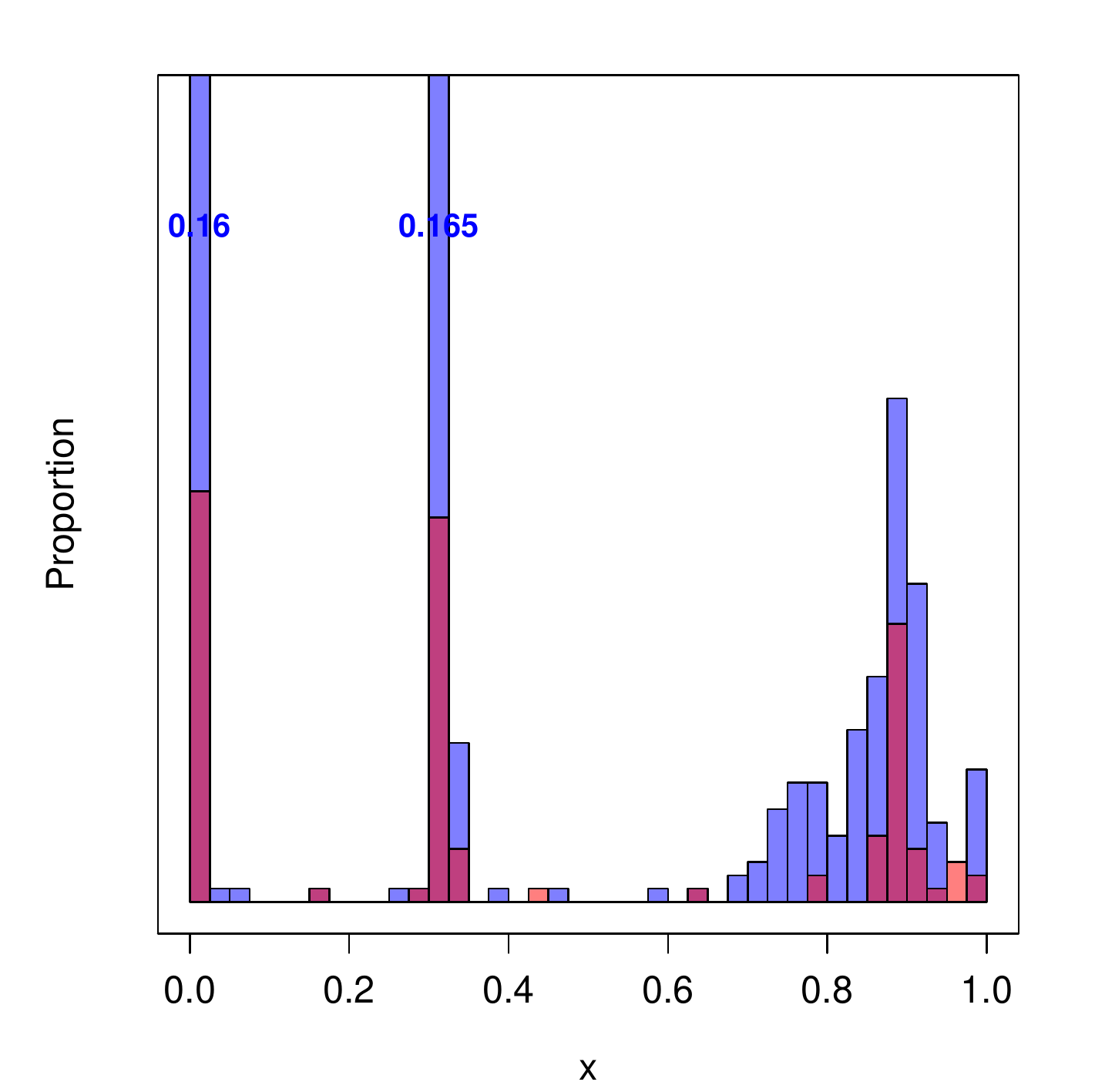}
 \includegraphics[width=0.48\textwidth,height=2.25in]{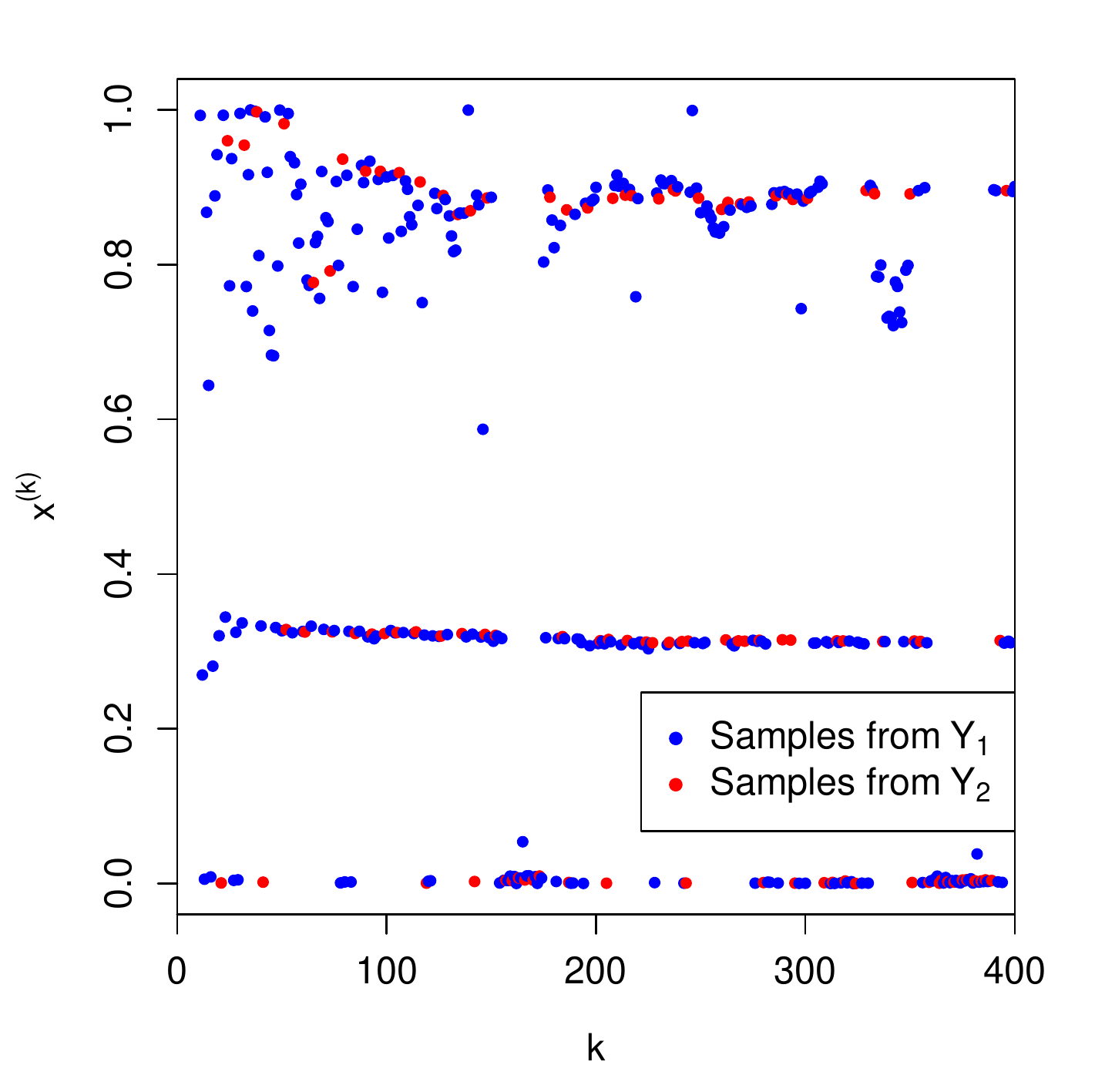}
  \caption{Left: the design $\ZZ^{(400)}$ based on the Gap-SUR EI criterion of \eqref{eq:MGap}. There were $D_1(400) = 294$ and $D_2(400)=106$ samples from $Y_1$ and $Y_2$ respectively. Right: sampled locations $x^k$ as a function of $k$ (blue for $\ell^k = 1$,  red for $\ell^k = 2$). \label{fig:1d} }
\end{figure}

From a different angle, Figure \ref{fig:1d} shows the resulting design $\ZZ^{(400)}$ in this example and the location of sampled sites $x^k$ as a function of sampling order $k=1,\ldots, 400$. We observe that the algorithm first engages in exploration and then settles into a more targeted mode, alternating between sampling around $0$, $r_1$ and $r_2$.



\subsection{Comparison and Discussion of EI Criteria}\label{sec:alterate-ei}

\red{As a first basis for comparison, we provide three non-adaptive designs.} The simplest alternative is the uniform sampling method that relies purely on the law of large numbers to learn $\mu_\ell(x)$. Thus, at each step $k$, we generate a new sampling location $(x,\ell)^k$ uniformly from $\XX \times \LL$. This generates a roughly equal number of samples $D_1(k)\simeq D_2(k)$ from each response and a kriging variance $\delta^2_\ell(x)$ that is approximately constant in $x$. Clearly, this approach yields an upper bound on the possible (empirical) loss. \red{ The second alternative is separate non-sequential modeling of each $\mu_\ell$ through a space-filling design (implemented via LHS); this improves on uniform sampling but does not attempt in any way to discriminate in the index dimension $\mk{L}$. For this example, we take $D_1 = 160 = 4 D_2$ to be proportional to the observation noise of each surface. (Note that this strategy  is roughly equivalent to building a global sequential maximin design using the acquisition function $E_k(x, \ell) := \delta_\ell(x)$.)}

\red{The third alternative is to build a sampling scheme that relies on the true $\mu_\ell(\cdot)$. With this foresight, we generate a design that relies on the \emph{actual} complexity for resolving $\CC(x)$ by plugging-in the true $\Delta_\ell(x)$ into the Gap-UCB metric in \eqref{eq:EGapSd}. Because sampling depends solely on $\Delta_\ell(x)$ and the kriging variances $\delta^{(k)}_\ell(x)^2$ are iteratively determined by the previous $x^{1:k}$, cf.~\eqref{eq:krig-cov}, the overall design $x^{1:K}$ is deterministic (hence non-adaptive, but still implemented sequentially). Note that the resulting $\widehat{\mu_\ell}(\cdot)$'s and hence outputted $\hat{\CC}(\cdot)$ are still a function of $Y^{1:K}$.}

%
%


Several further alternatives for evaluating expected improvement can be designed based on classification frameworks. For classification, the main posterior statistic is the probabilities $p_\ell(x)$ of $\mu_\ell(x)$ being the smallest response. One can then use the vector $\vec{p}(x)$ to measure the complexity of the resulting local classification at $x$. Note that such measures intrinsically aggregate across $\ell$ and hence only depend on $x$. This suggests either using a two-step sampling procedure as in Section \ref{sec:two-step} or building a UCB-like criterion as in \eqref{eq:EGapSd}. \red{We employ the latter method, blending a criterion $\Gamma(x)$ that discriminates among $x$-locations (larger scores are preferred) with UCB, leading to EI scores of the form $E_k(x,\ell) = \Gamma^{(k)}(x) + \gamma_k \delta_\ell(x)$. }
Three different choices for $\Gamma(\cdot)$ are:
 \begin{align}
 \Gamma^{ENT}(x) &:= -\sum_\ell p_\ell(x)\log p_\ell(x);\label{eq:P-entropy} \\
  \Gamma^{BvSB}(x) &:= -\left[p_{Best}(x) - p_{SB}(x)\right];  \label{eq:BvSB}  \\
  \Gamma^{Best}(x) & := -p_{Best}(x) \label{eq:best},
 \end{align}
where $p_{Best}(x) := \PP\left(\hat{\CC}(x) = \CC(x) | \FF_k\right)=p_{\hat{\CC}(x)}$ is the posterior probability that the lowest posterior mean is indeed the smallest response, and $p_{SB}$ is the probability that the second-lowest posterior mean is the smallest response.

\red{The $\Gamma^{ENT}$ metric is the posterior entropy which is a standard measure of classification complexity.
High entropy indicates more spread in $\vec{p}(x)$ and hence more uncertainty about which is the smallest component of $\vec{\mu}(x)$. However, a well-known drawback of entropy is that for large $L$ (bigger than 3) the responses that are very unlikely to be the minimum (i.e.~small $p_\ell(x)$) still strongly affect the overall $\Gamma^{ENT}(x)$, leading to non-intuitive shapes of the EI scores. The Best-versus-Second-Best (BvSB) approach $\Gamma^{BvSB}(x)$ originating in \cite{JoPoPa:09}, counteracts this effect by comparing just the two lowest posterior means. Small differences between $p_{Best}$ and $p_{SB}$ indicate large uncertainty in identifying the minimum response.  The BvSB metric can break down however if posterior variances $\delta_\ell(x)$'s are highly unequal, whereby the ordering between $\widehat{\mu_\ell}$ and $p_\ell$'s is not the same. Otherwise, $\Gamma^{BvSB}$ is quite similar to the gap measure $\widehat\Delta(x)$. Lastly, $\Gamma^{Best}$ focuses on the locations where $p_{Best}(x) \ll 1$, i.e.~those close to classification boundaries of $\hat\CC(x)$. When $L=2$, $\Gamma^{Best}$ and $\Gamma^{BvSB}=1-2p_{Best}(x)$ give the same preferences.}

\red{Note that because $\Gamma$ does not discriminate among the surfaces, it is sensible to take $\gamma_k = \gamma_k(\ell)$ to be response-specific. Alternatively, the $\Gamma$ metrics lend themselves to concurrent sampling which builds an adaptive sequential design in $\XX$ but treats all surfaces equally:
\begin{align}\label{eq:conc-best}
E^{Conc-\Gamma}_k(x) = \Gamma^{(k)}(x) + \gamma_k [\sum_\ell \delta_\ell(x)].\end{align}}

Yet another alternative is a so-called pure M-Gap heuristic that uses \eqref{eq:MGap} via
\red{\begin{align}\label{eq:pure-Mgap} x^{k+1} =\arg \max_{x \in \TT^{(k)}} \cM(x), \qquad \ell^{k+1} = \arg\max_\ell \delta^2_\ell(x^{k+1}).
  \end{align}} This hierarchical sampling strategy can be viewed as generalizing the Efficient Global Optimization (EGO) criterion of \cite{JonesSchonlauWelch98} to the ranking problem, cf.~the classification variant of EGO in \cite{tgpPackage}. 



\subsection{Benchmarks}\label{sec:benchmark}

\begin{table}[htb]
\centering
\caption{True loss \emph{vs.}~empirical loss with $\ZZ^{(200)}$ for the 1-D example. For UCB heuristics the cooling schedule is of the form $\gamma_k = c \sqrt{\log{k}}$ with $c$ as listed below. The error probability $ErrProb$ measures the mean of $1-p^{(200)}_{Best}(x)$ over the test set. $D_1 = D_1(200)$ is the number of samples out of 200 total from $Y_1$.\label{tbl:benchmark}}
{\small \begin{tabular}{lccccccc}
  \hline
  {Method} & {Emp Loss} & (SE) & {True Loss} & {(SE)} & {ErrProb} & (SE) & $D_1$ \\ \hline\hline
  {Uniform Sampling} & 2.89E-3 & (1.24E-4) & 2.64E-3  & (2.67E-4) & 6.87\% & (0.25\%) & 100 \\
  {Non-adaptive LHS} &  2.16E-3  &  (1.01E-4)  &  1.91E-3  &  (2.12E-4)  &  6.05\%  &  (0.22\%)  & 160 \\
   {Known-Gap-UCB, $c=4$ } & 1.77E-3 & (8.35E-5) & 1.43E-3 & (1.91E-4) & 5.61\% & (0.23\%) & 174 \\
  {Gap-SUR} &  0.96E-3 & (4.98E-5) & 1.19E-3 & (1.84E-4) & 3.82\% & (0.17\%) & 146 \\
  {Pure M-Gap} & 1.20E-3 & (5.39E-5) & 1.81E-3 & (2.33E-4) & 4.28\% & (0.15\%) & 172 \\
  {Concurrent M-Gap} & 1.36E-3 & (8.33E-5) & 1.52E-3 & (1.97E-4) & 4.78\% & (0.24\%) & 100 \\ \hline
   {Gap-UCB}, $c = 0.1$& 2.62E-3 & (1.74E-4) & 2.23E-3  & (2.60E-4) & 5.46\% & (0.23\%) & 163 \\
			{Gap-UCB}, $c = 0.25$& 2.05E-3 & (1.02E-4) & 1.63E-3  & (2.43E-4) & 5.16\% & (0.19\%) & 165 \\
 {Gap-UCB}, $c = 1$  & 1.27E-3 &(5.61E-5) & 1.50E-3 & (1.98E-4) & 4.39\% & (0.16\%) & 167 \\			
			{Gap-UCB}, $c = 5$& 1.56E-3 & (7.29E-5) & 1.62E-3  & (2.14E-4) & 5.10\% & (0.20\%) & 176 \\
			{Gap-UCB}, $c = 10$& 1.83E-3 & (7.89E-5) & 1.48E-3  & (1.89E-4) & 5.49\% & (0.20\%) & 172 \\
  {$\Gamma^{Best}$-UCB}, $c = 5$& 1.29E-3 & (5.85E-5) & 1.35E-3  & (1.71E-4) & 4.53\% & (0.17\%) & 172 \\
			{$\Gamma^{ENT}$-UCB}, $c = 5$& 1.14E-3 & (6.02E-5) & 1.33E-3  & (1.80E-4) & 4.22\% & (0.18\%) & 169 \\
  \hline
    {Gap-SUR w/training $\cK_1$} & 1.20E-3 & (5.87E-5) & 1.69E-3 &  (3.24E-4) &   4.34\% & (0.37\%) & 146 \\
\hline \hline
\end{tabular}}
\end{table}

\begin{figure}[htb]
\centering
\includegraphics[height=2.6in,width=0.49\textwidth]{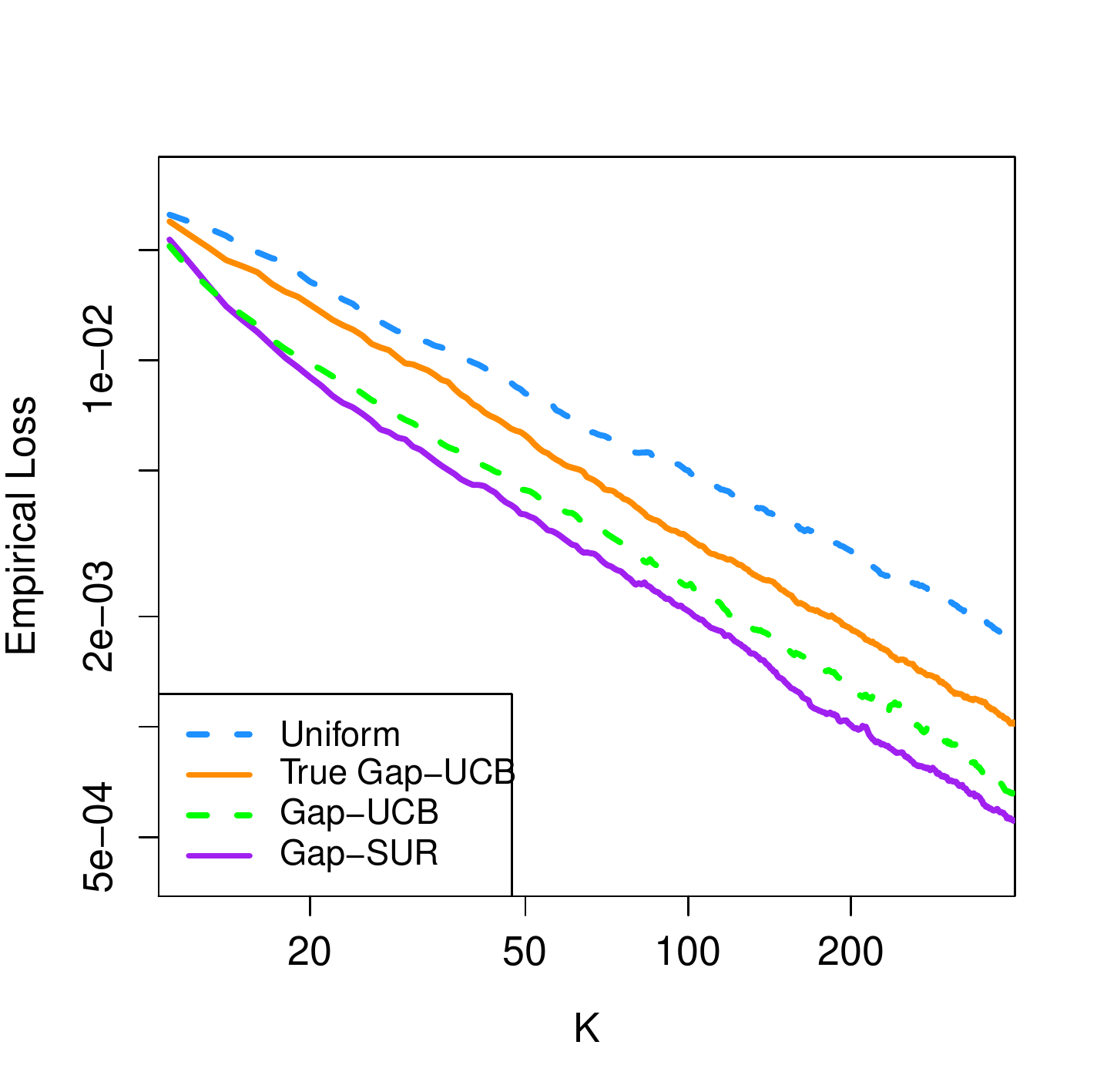}  
\includegraphics[height=2.6in,width=0.49\textwidth]{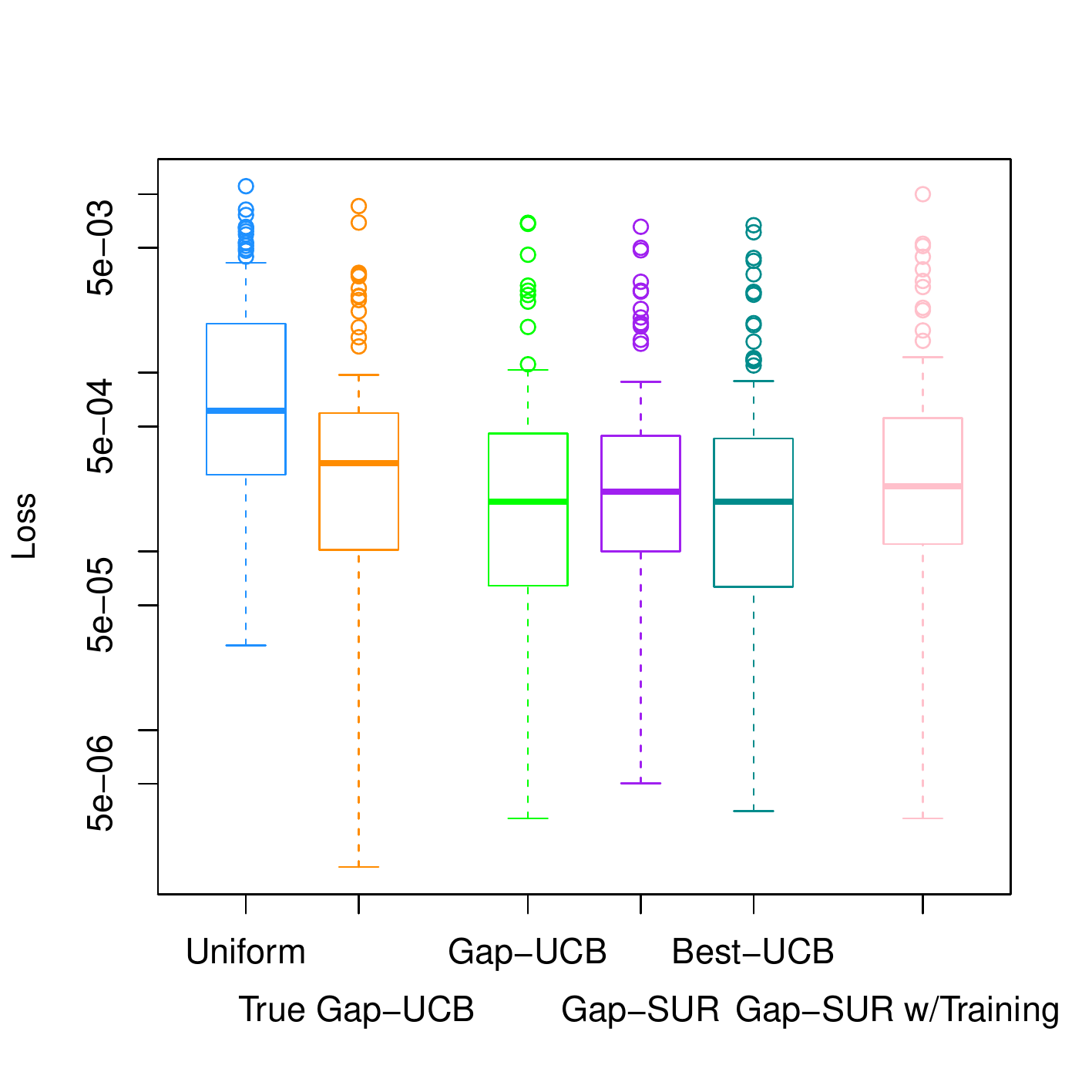}
\caption{Left: Averaged empirical loss $\mc{EL}(\hat{\CC}^{(k)})$ as a function of design size $k$ (in log-log scale). We compare our adaptive Gap-SUR \eqref{eq:MGap} and Gap-UCB  methods \eqref{eq:EGapSd} (with $\gamma_k = 1\cdot \sqrt{\log k}$) against a uniform sampler and a Known-Gap-UCB based on the true gap $\Delta(\cdot)$. Right: boxplot of $\LL(\hat\CC^{(K)},\CC)$ at $K=400$ computed via \eqref{eq:emp-el}, across six different EI approaches.
\label{fig:ave-emp-loss}}
\end{figure}

\red{To judge the efficiency of different sequential designs, we proceed to benchmark the performance of different approaches. Table \ref{tbl:benchmark} and Figure \ref{fig:ave-emp-loss}
compare the performance of  EI acquisition functions, including the three non-adaptive methods; Gap-SUR; Gap-UCB with different $\gamma_k$-schedules; methods based on posterior probabilities $\vec{p}(\cdot)$:
$\Gamma^{ENT}$-UCB entropy criterion based on \eqref{eq:P-entropy} and $\Gamma^{Best}$-UCB criterion based on \eqref{eq:best}; the pure M-gap heuristic \eqref{eq:pure-Mgap}; and concurrent sampling with M-Gap.}
To construct the summary statistics in Table \ref{tbl:benchmark} we initialized each algorithm with a random LHS design of size $K_0 =10$ and augmenting it until $K=200$ sites. Throughout, we compute both the true loss in this synthetic example where $\mu_\ell(x)$ are known, as well as the approximated empirical loss $\mc{EL}$
\begin{align}\label{eq:emp-el}
\mc{EL}(\hat\CC,\CC) &= \frac{1}{M}\sum_{j=1}^M \left\{ \widehat\mu_{(1)}(j\Delta x) - m(j\Delta x)\right\},
\end{align}
where we used $M=1000 = 1/\Delta x$ uniformly spaced gridpoints in $\XX=[0,1]$. A further metric reported is the error probability $1-p_{Best}^{(K)}(x)$ which measures the posterior probability that the identified minimum response is incorrect. \red{Each method was run 100 times to compute the resulting mean and standard deviation of the loss function $\LL$ and the empirical loss $\mc{EL}$.} To isolate the effect of the EI criterion, we continue with a fixed GP covariance structure $\cK_\ell$ for the $\mu_\ell$'s \red{and pre-specified $\sigma_\ell$'s} (see hyperparameter values in Sec.~\ref{sec:1d}).

The Gap-SUR algorithm appears to be the most efficient, in particular being much more efficient than a naive uniform sampler (or the non-adaptive LHS sampler). It also performs  better than Gap-UCB or the pure M-Gap methods and moreover also has the smallest fluctuations across algorithm runs, indicating more stable behavior. Nevertheless, the UCB methods are nearly as good, in particular the entropy-based $\Gamma^{ENT}$-UCB approach is competitive. \red{However, as discussed these methods are sensitive to the choice of the $\gamma_k$-schedule; the table shows that a poorly chosen $\gamma_k$ can materially worsen performance. In this example, with $\gamma = c\sqrt{ \log k}$, the scaling $c=1$ works well, but if $c$ is too small then the method is overly aggressive, and if $c$ is too big the sampling is essentially space-filling.} At the same time, a limitation of Gap-SUR is that it requires knowing the noise variances $\sigma^2_\ell(\cdot)$ when optimizing the EI acquisition function.  \red{ Perhaps surprisingly, the Known-Gap-UCB strategy loses out to the adaptive methods. This happens because the empirical loss of the non-adaptive method is in fact rather sensitive to the observed samples $Y^{1:K}$ which can generate erroneous estimates of $\mu_\ell(x)$ and mis-classified $\CC(x)$. Consequently the Known-Gap-UCB design, while properly placing $(x,\ell)^{1:K}$ \emph{on average}, does not allow self-correction so that erroneous beliefs about $\mu_\ell$ can persist for a long time, increasing $\mc{EL}$. In contrast, adaptive algorithms add samples to any regions where observations suggest that $\Delta(x)$ is small, sharpening accuracy there and lowering both true and empirical loss functions.}

The left panel of Figure \ref{fig:ave-emp-loss} visualizes algorithm behavior as a function of design size $k$, by plotting the approximated empirical loss $\mc{EL}(\hat\CC^{(k)},\CC)$ from \eqref{eq:emp-el} for
 four representative strategies. All methods appear to enjoy a power-law (linear behavior on the log-log plot) for $\mc{EL}$ as a function of $k$, with the slopes of the adaptive method strictly bigger than the non-adaptive ones.  


Table \ref{tbl:benchmark} also highlights the gain from discriminating among the response surfaces, as the Concurrent M-Gap algorithm is notably worse (with losses of about 30\% higher) relative to Gap-SUR. The only difference between these methods is that Gap-SUR sampled $Y_1$ 146 times out of 200, while the concurrent method was constrained to sample each response exactly 100 times. All approaches that optimize over the full $\XX \times \mk{L}$ focus on fitting the noisier $Y_1$, sampling it 70--85\% of the rounds (see the $D_1$ column).

As a final comparison, the last row of Table \ref{tbl:benchmark} reports the performance of the Gap-SUR method in the practical context where one must also \emph{train} the GP kernels $\cK_\ell$'s by learning $\theta_i, s^2, \sigma$. \red{All the parameters, including the observation noise $\sigma$ which is viewed as the nugget of the GP covariance structure, are estimated via MLE.
 Since training introduces additional noise into the fitted response surfaces, algorithm performance is necessarily degraded, especially in terms of variation across algorithm runs. This could indicate that the stationary GP model is not ideal here.}

Table \ref{tbl:benchmark} also shows that the empirical $\mc{EL}(\hat\CC^{(K)})$ and actual loss $\LL(\hat\CC^{(K)},\CC)$ metrics are consistent, so that the former can be used as an internal online assessment tool to monitor accuracy of the estimated classifier. Mismatch between the two measures is driven by model mis-specification, as incorrectly inferred covariance structure of $\mu_1(x)$ leads to over-optimism: $\mc{EL} < \LL$. This issue is largely independent of the sampling scheme and pertains more to the modeling framework than to EI acquisition functions.

\subsection{Many Surfaces}

\begin{figure}[ht]
  \centering
  \begin{tabular}{ccc} \hspace*{-0.25in}
  \includegraphics[width=0.32\textwidth,height=2in,trim=0.1in 0.1in 0.1in 0.1in]{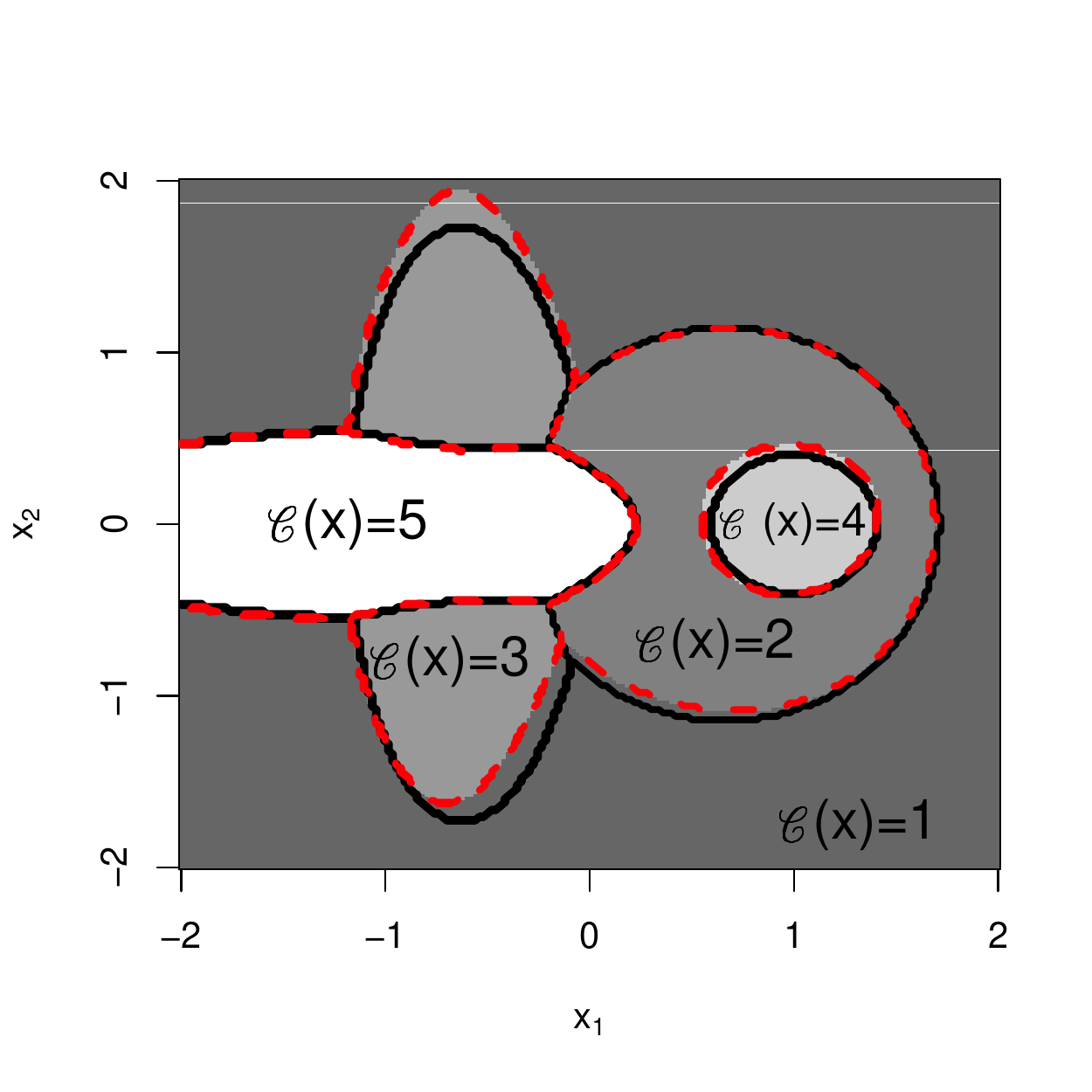} & \includegraphics[width=0.32\textwidth,height=2in,trim=0.1in 0.1in 0.1in 0.1in]{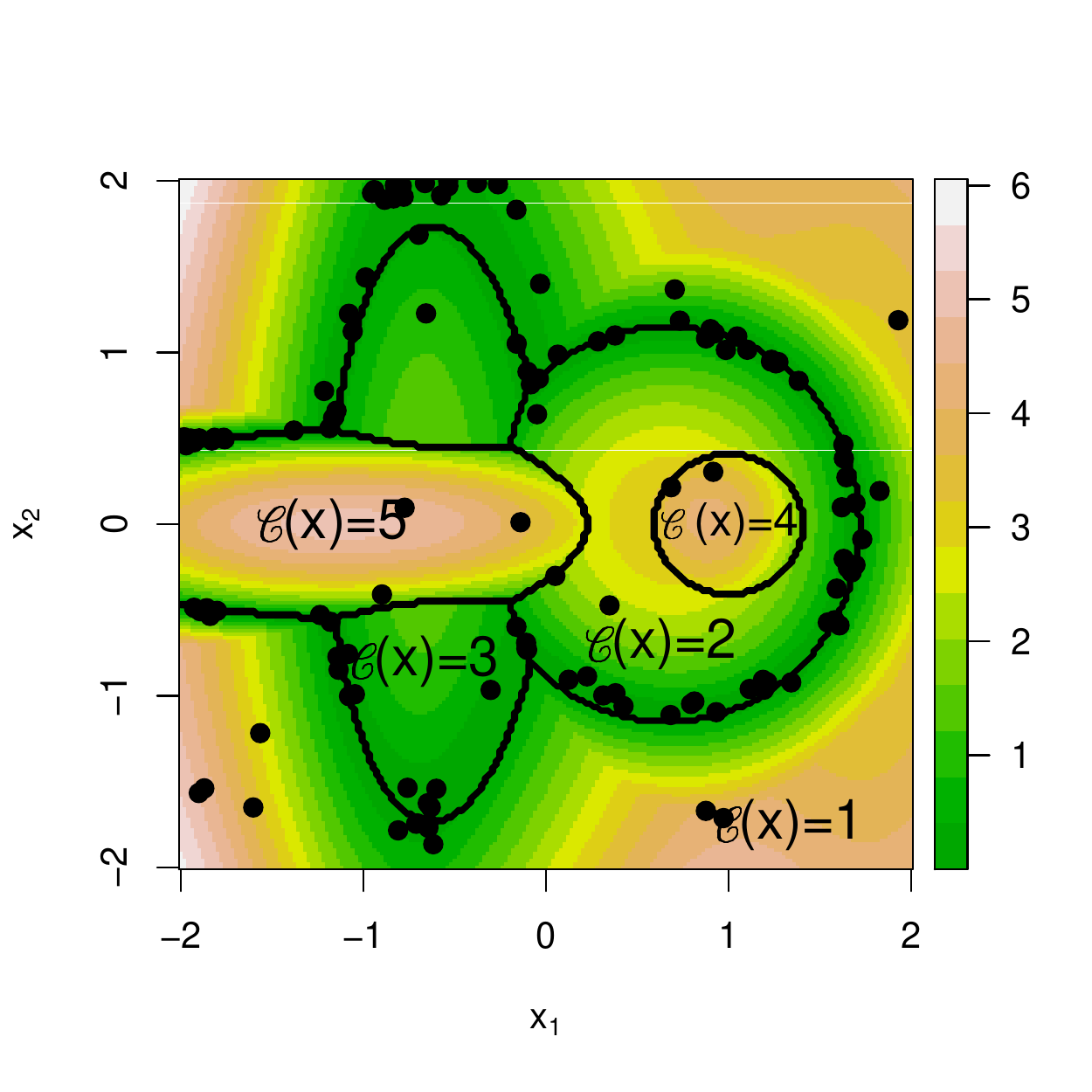} &
  \includegraphics[width=0.32\textwidth,height=2in,trim=0.1in 0.1in 0.1in 0.1in]{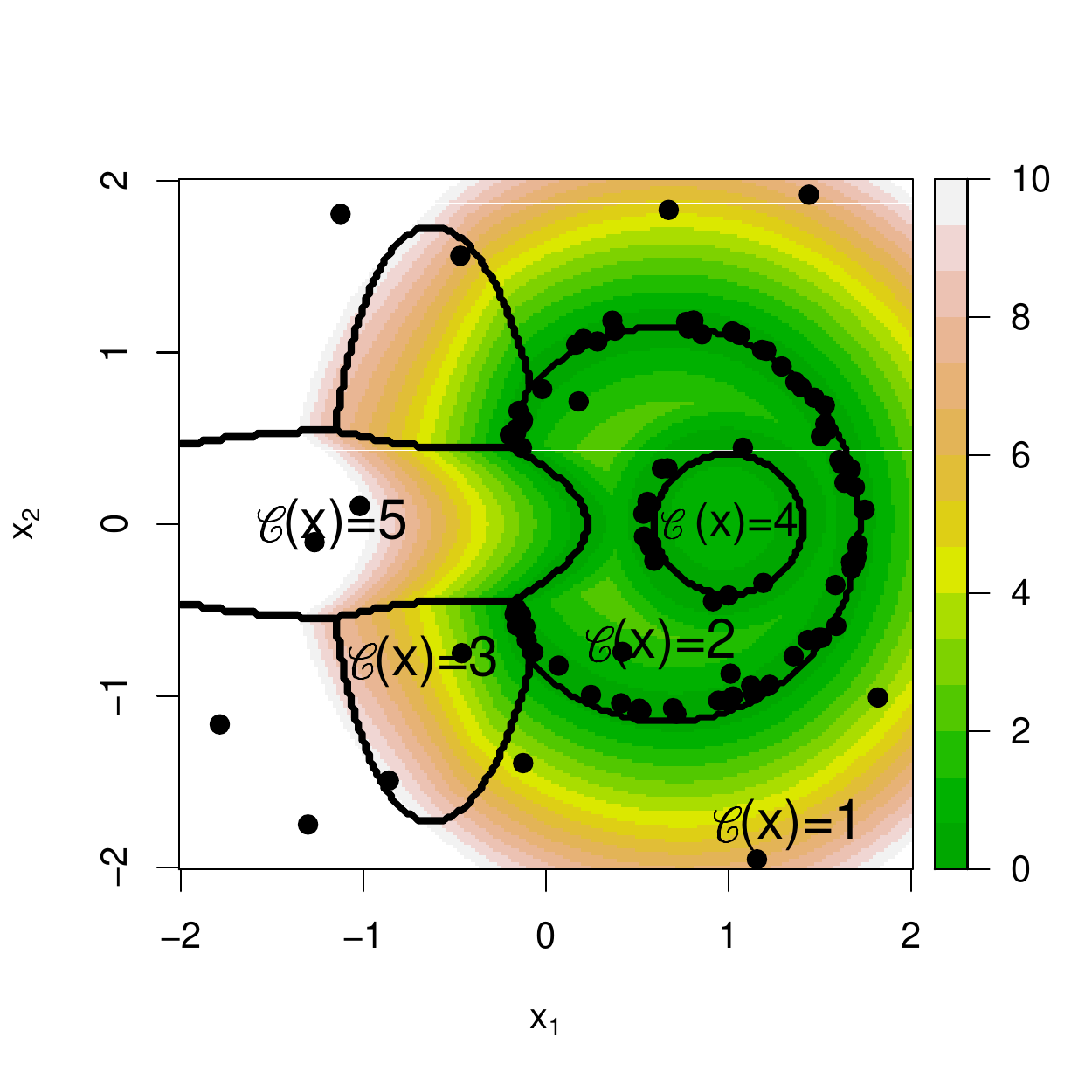} \\
   Overall $\hat{\CC}$ & $\ell=1$ & $\ell=2$  \\
    \includegraphics[width=0.32\textwidth,height=2in,trim=0.1in 0.1in 0.1in 0.1in]{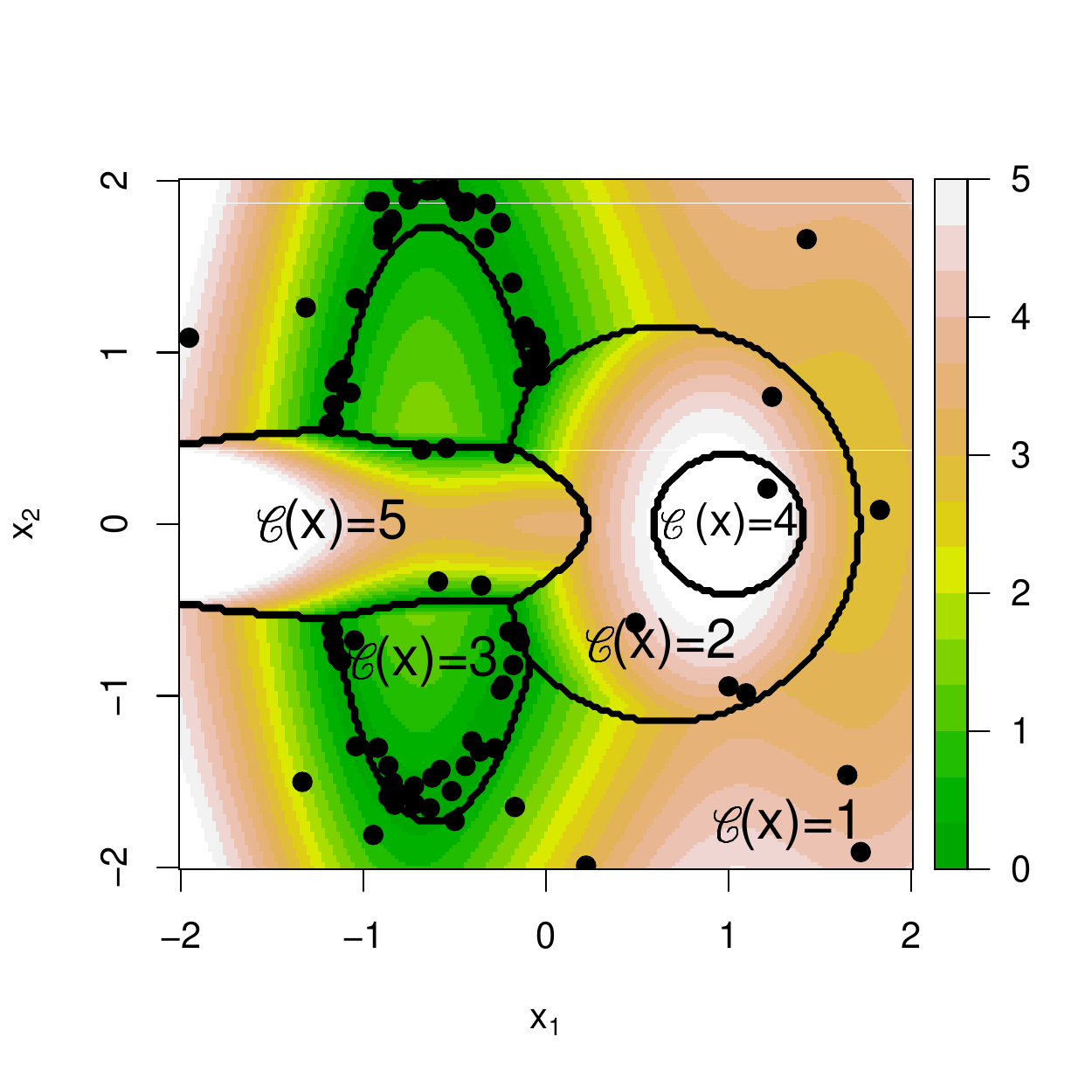} &
  \includegraphics[width=0.32\textwidth,height=2in,trim=0.1in 0.1in 0.1in 0.1in]{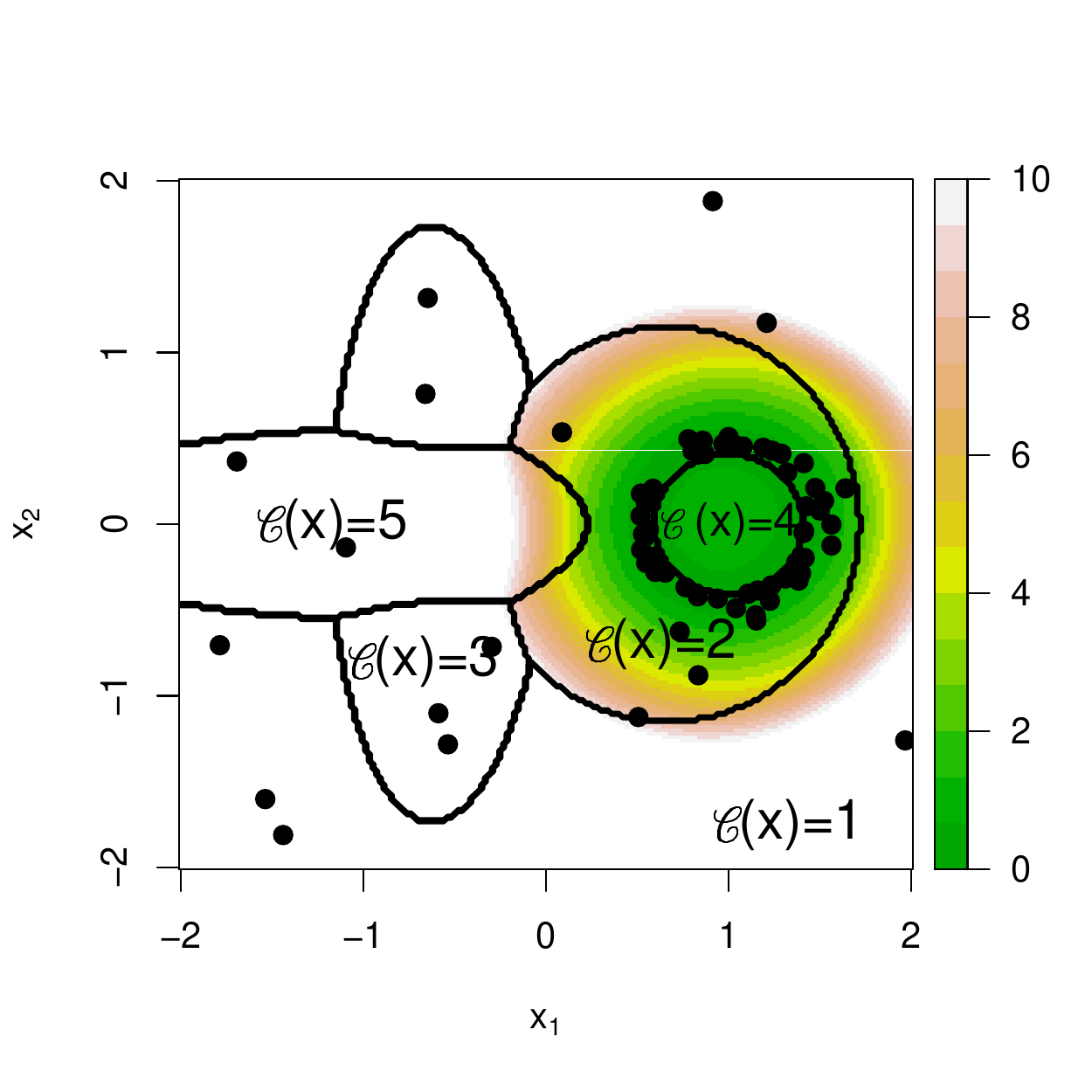} &
   \includegraphics[width=0.32\textwidth,height=2in,trim=0.1in 0.1in 0.1in 0.1in]{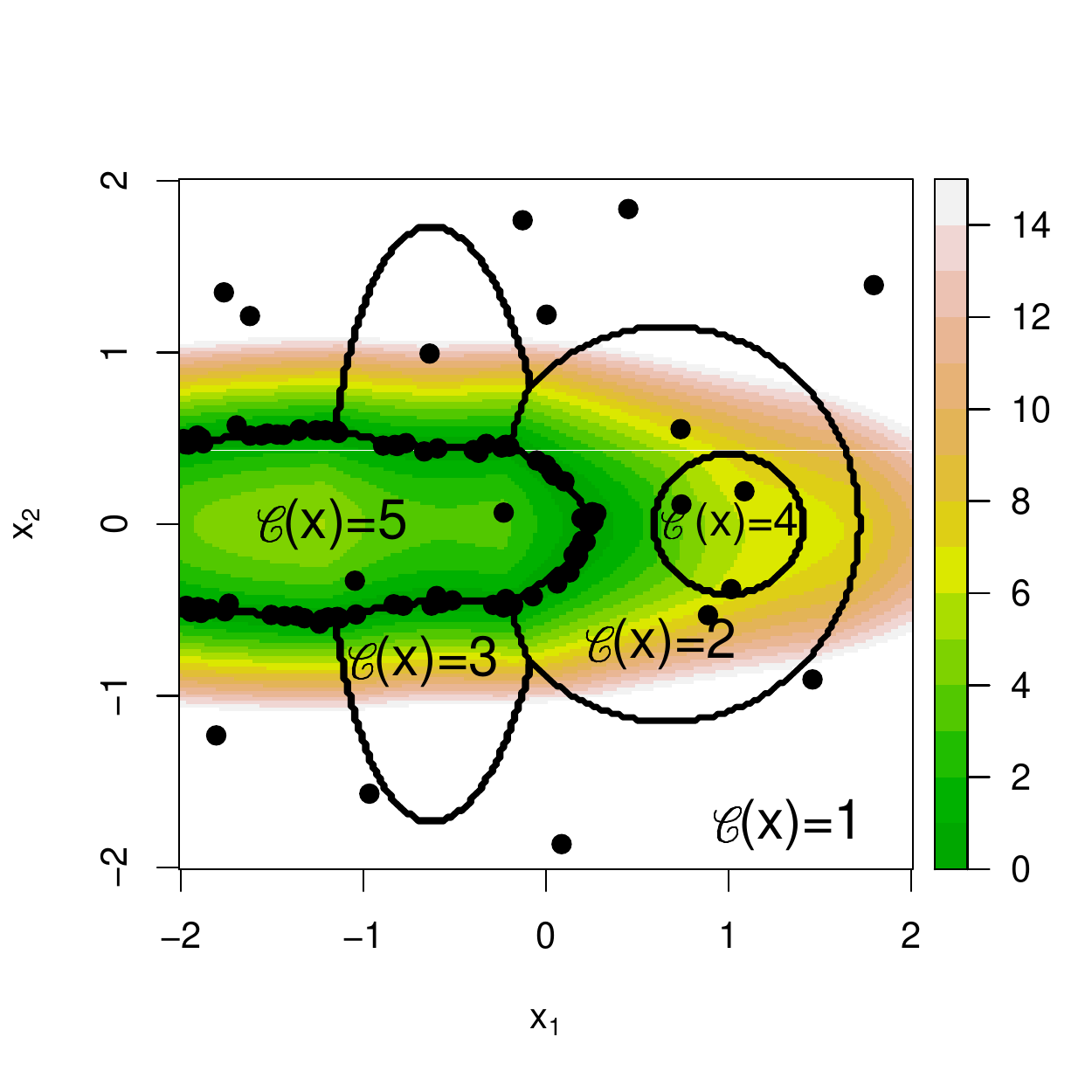}  \\
  $\ell=3$ & $\ell=4$ & $\ell=5$  \end{tabular}
  \caption{\red{2-D Ranking on $\XX = [-2,2] \times [-2,2]$ using the Gap-SUR heuristic. \emph{Top-left} panel: The solid black lines show the true $\CC(x_1,x_2)$, the dashed red lines show the estimated classifier $\hat\CC^{(K)}(x_1,x_2)$ for $K=500$. The other panels show the marginal designs $(x_1,x_2)^{1:D_\ell(K)}$ for each of the 5 response surfaces. Shading indicates the estimated empirical gaps $\widehat\Delta_\ell(x_1,x_2)$, $\ell = 1, \ldots, 5$. We observe that most samples gravitate towards regions where $\widehat\Delta_\ell \simeq 0$. Solid curves indicate boundaries of the true classifier $\CC(x_1,x_2)$.} \label{fig:2d}}
\end{figure}

\red{Our next example treats a more complex setting with $L=5$ surfaces and a 2-dimensional input space $\XX=[-2,2]^2$:
$$\begin{array}{lrr}
\hline
  & \text{Response} & \text{GP Parameters }  (\theta_1, \theta_2, \eta^2, t_\ell) \\ \hline\hline
\mu_1(x_1,x_2) &  2-x_1^2 - 0.5 x_2^2 & (4,6.5,23,-10)\\
\mu_2(x_1,x_2) &  2(x_1-1)^2 + 2x_2^2 -2 & (7.5,7.5,475,60)\\
\mu_3(x_1,x_2) &  2 \sin(2x_1)+2 &  (1,8,2,1.9) \\
\mu_4(x_1,x_2) &  8(x_1-1)^2 + 8x_2^2 -3 & (8,8,8000,300) \\
\mu_5(x_1,x_2) &  0.5(x_1+3)^2 +16x_2^2 -6 & (8,4,2500,150) \\ \hline
\end{array}$$
We assume constant homoskedastic observation noise $\eps_\ell(x_1,x_2) \sim \mathcal{N}(0, \sigma_\ell^2)$,  $\sigma_\ell=0.5 \; \forall \ell$. The GP models have separable anisotropic Matern-5/2 covariance functions with the specified hyperparameters, and fixed trend $t_\ell$. Figure \ref{fig:2d} shows the corresponding  classifier $\CC$.}

The sequential designs were initialized at $K_0 =50$ by generating 10 LHS samples from each $Y_\ell(x_1, x_2)$;
at each step the sampling locations were selected from a LHS candidate set $\TT$ of size $D=100$ using the randomized $\eps$-greedy method with $\eps=0.1$. 



\red{The top-left panel of Figure \ref{fig:2d} shows the estimated classifier $\hat\CC^{(K)}$ after $K=500$ samples in total using the Gap-SUR acquisition function, and the other panels display the locations of the sampled $x$'s as allocated for each $\ell=1,\ldots,5$. As can be seen, the algorithm is highly discriminating in sampling jointly on $\XX \times \LL$. At any given classification boundary, the algorithm effectively only sampled two out of the five responses, endogenously recovering the concept of Best-versus-Second Best testing.  Thus, samples from $Y_\ell$ are mostly located around the boundaries of surface $\mu_\ell$ and other surfaces. These contours, where $\Delta_\ell = \mu_\ell - \min_{j \neq \ell} \mu_j  =0$, are precisely the regions targeted by the Gap EI metrics. Because $\CC_1$ and $\CC_5$ have the longest boundaries, relatively more samples were  chosen there ($D_1=126, D_5=109$); conversely the smallest set is $\CC_4$ which only received $D_4 = 70$ samples.}

\red{Table \ref{tbl:2d} presents the relative performance of different acquisition functions. Specifically, we compare (i) uniform sampling; (ii) space-filling LHS sampling; (iii) concurrent $\Gamma^{Best}$ strategy \eqref{eq:conc-best} which is analogous to entropy-based sampling; (iv) Gap-UCB, and (v) Gap-SUR. We note that with many surfaces, the key is not necessarily the budget allocation among the surfaces (here, with identical $\sigma_\ell$, optimal $D_\ell$'s are roughly equal), but efficient placement of sample locations that are most appropriate for each surface. This effect can be observed by comparing a non-adaptive strategy (that is space-filling in both $x$ and $\ell$), to a concurrent $\Gamma^{Best}$ strategy \eqref{eq:conc-best} (that targets classification boundaries but is uniform in $\ell$), to a Gap-SUR/Gap-UCB strategy (that targets different parts of classification boundaries for different indices $\ell$). Each step in the above sequence generates substantial performance gains; it is expected to be even more pronounced when the observation noise is index- (or state-) dependent.}


\begin{table}[htb]
\centering
\red{\caption{True loss \emph{vs.}~empirical loss with $\ZZ^{(500)}$ for the 2-D example. For UCB heuristics the cooling schedule is of the form $\gamma_k = c \sqrt{\log{k}}$. The error probability is $ErrProb= Ave( 1-p^{(500)}_{Best}(x))$ over the test set. The vector $D_\ell(500)$ lists the number of samples out of 500 total from $Y_\ell$, $\ell=1,\ldots,5$.\label{tbl:2d}}	
	{\small \begin{tabular}{lccccccc}
			\hline
			{Method} & {Emp Loss} & (SE) & {True Loss} & {(SE)} & {ErrProb} &   Index Allocations $D_\ell$\\ \hline\hline
			{Uniform Sampling}& 6.43E-3 & (4.64E-5) & 5.47E-3  & (2.39E-4) & 4.10\% & 
(100,100,100,100,100)\\
			{Non-Adaptive LHS} & 5.97E-3 & (2.31E-5) & 4.72E-3  & (1.97E-4) & 3.92\% & 
(100,100,100,100,100)\\
			{Conc $\Gamma^{Best}$, $c = 0.5$} & 5.11E-3 & (1.93E-5) & 4.04E-3  & (1.50E-4) & 3.66\% & 
(100,100,100,100,100)\\
			{Gap-SUR} &  3.46E-3 & (1.32E-5) & 3.17E-3 & (1.29E-4) & 3.06\% & 
(126, 101, 94, 70, 109) \\
			{Gap-UCB}, $c = 0.5$ & 3.41E-3 & (1.45E-5) & 2.97E-3& (1.14E-4) & 3.05\% & 
(129, 103, 104, 72, 92) \\
			\hline \hline
		\end{tabular}}}
\end{table}

\section{Case Study in Epidemics Management}\label{sec:epi}

Our last example is based on control problems in the context of infectious epidemics \cite{LudkovskiLin14,LN10,LN11wsc,MerlGramacy09}. Consider the stochastic SIR model which is a compartmental state-space model that partitions a population pool into the three classes of Susceptible counts $S_t$, Infecteds $I_t$ and Recovereds $R_t$. We assume a fixed population size $M = S_t + I_t + R_t$ so that the state space is the two-dimensional simplex $\XX = \{(s,i) \in \mathbb{Z}^2_+ : s+ i \le M \}$. In a typical setting, $M \in [10^3, 10^5]$, so that $\XX$ is discrete but too large to be explicitly enumerated (on the order of $| \XX | \simeq 10^6$).
The dynamics of $(S_t, I_t)$ are time-stationary and will be specified below in \eqref{eq:sir}.

The goal of the controller is to mitigate epidemic impact through timely intervention, such as
%
social distancing measures that lower the infectivity rate by reducing individuals' contact rates; mathematically this corresponds to modifying the dynamics of $(S_t, I_t)$. To conduct cost-benefit optimization, we introduce on the one hand epidemic costs, here taken to be proportional to the number of cumulative infecteds, and on the other hand intervention costs, that are proportional to the current number of remaining susceptibles   $C^I S_t$. Intervention protocol can then be (myopically) optimized by comparing the expected cost of no-action $\mu_0(s,i)$ (conditional on the present state $(s,i)$) against the expected cost of immediate action, $\mu_A(s,i)$. More precisely, let
\begin{align}\label{eq:epi-cost-1}
\mu_{0}(s,i) &:= \EE^{0}[ S_0 - S_T | I_0 = i, S_0 = s] \quad \text{and}\\ \label{eq:epi-cost-2}
\mu_{A}(s,i) &:= \EE^{A}[ S_0  -S_T  | I_0 = i, S_0 = s] + C^I s.
\end{align}
Above, $T = \inf\{ t: I_t = 0\}$ is the random end date of the outbreak; due to the fixed population and posited immunity from disease after being infected, the epidemic is guaranteed to have a finite lifetime. The difference $S_0-S_T$ thus precisely measures the total number of original susceptibles who got infected at some point during the outbreak.

The overall goal is then to \emph{rank} $\mu_{0}$ and $\mu_A$, with the intervention region corresponding to $\{ (s,i) : \mu_A(s,i) > \mu_{0}(s,i) \}$. Because no analytic formulas are available for $\mu_\ell$'s, a  sensible procedure (also preferred due to the ease of handling numerous extensions of SIR models) is a Monte Carlo sampler that given an initial condition $S_0=s,I_0=i$ and regime $\ell \in \{0, A\}$ generates a trajectory $(S_t, I_t)(\omega)$ and uses it to evaluate the pathwise $S_T(\omega)$, connecting to the framework of \eqref{def_cal}.

From the policy perspective, the trade-off in \eqref{eq:epi-cost-1}-\eqref{eq:epi-cost-2} revolves around doing nothing and letting the outbreak run its course, which carries a unit cost for each individual that is eventually infected, or implementing preventive social distancing measures which costs $C^I$ for each \emph{susceptible}, but lowers the expected number of future infecteds. Typical countermeasures might be public ad campaigns, school closures, or distribution of prophylactic agents. In general, intervention is needed as soon as there is a threat of a big enough outbreak. However, if $I_t$ is low, the cost of intervention is too high relative to its benefit because the epidemic might end on its own. Similarly, if $S_t$ is low, the susceptible pool is naturally exhausted, again making intervention irrelevant (due to being ``too late''). Quantifying these scenarios requires a precise probabilistic model.

The dynamics of $(S_t, I_t)$ under the respective laws $\PP^{0}$ and $\PP^{A}$ follow continuous-time Markov chains with the following two transition channels:
\begin{align}\label{eq:sir}
\left\{ \begin{aligned}
\text{Infection}: & S+I \to 2I  & &\text{with rate}\;\; \beta^{j} S_t I_t/M, \quad j=0,A; \\
\text{Recovery}: & I \to R  & &\text{with rate}\;\; \gamma I_t. \\ \end{aligned}\right\}
\end{align}
Above, $\beta^{A} < \beta^0$ is interpreted as lowered contact rate among Infecteds and Susceptibles in the intervention regime, which thereby reduces outbreak growth and impact. The Markov chain $(S_t, I_t)$ described in \eqref{eq:sir} is readily simulatable using the Gillespie time-stepping algorithm \cite{Gill:exac:1977}, utilizing the fact that the sojourn times between state transitions have (state-dependent) Exponential distributions, and are independent of the next transition type. These simulations are however rather time-consuming, requiring $\OO(M)$ Uniform draws. Consequently, efficient ranking of expected costs is important in applications.

\begin{remark}
Since \eqref{eq:sir} implies that each individual infected period has an independent $Exp(\gamma)$ distribution it follows that
$
\EE[ S_0 - S_T ] = \gamma \EE \bigl[ \int_0^T I_t \,dt \bigr],
$
so that \eqref{eq:epi-cost-1} can also be interpreted as proportional to total expected infected-days.
\end{remark}

We note that in this example the input space $\XX$ is discrete, which however requires minimal changes to our implementation of Algorithm \ref{algorithm}. The biggest adjustment is the fact that the noise variances $\sigma^2_\ell(x)$ in \eqref{def_Y} are unknown. Knowledge of $\sigma^2_\ell(x)$'s is crucial for training the GP covariance kernel $\cK_\ell$, see e.g.~\eqref{eq:krig-mean}. Indeed, while it is possible to simultaneously train $\cK_\ell$ and a constant observation noise $\sigma$ (\red{the latter is known as the ``nugget'' in GP literature, and can be inferred via maximum likelihood}), with state-dependent noise $\cK$ is not identifiable.
 We resolve this issue through a batching procedure (compare to \cite[Sec 3.1]{NelsonStaum10}) to estimate $\sigma^2_\ell(x)$ on-the-go. 
 Namely, we re-use the same site $x \equiv (s,i)$ $r$-times,
to obtain independent samples $y^{(1)}_\ell(x), \ldots, y^{(r)}_\ell(x)$ from the corresponding $Y_\ell(x)$. This allows to estimate the conditional variance
$$
\widetilde{\sigma}^2_\ell(x) := \frac{1}{r-1} \sum_{i=1}^r (y^{(i)}(x) -\bar{y}_\ell(x))^2, \quad\text{ where  } \quad \bar{y}_\ell(x) = \frac{1}{r} \sum_{i=1}^r y^{(i)}_\ell(x)
 $$
 is the sample mean. Moreover,  as shown in \cite[Sec 4.4.2]{GinsbourgerPicheny13} we can treat the $r$ samples at $x$ as the single design entry $(x,\bar{y}_\ell(x))$ with noise variance $\widetilde{\sigma}^2_\ell(x)/r$.
 The resulting reduction in post-averaged design size by a factor of $r$ offers substantial computational speed-up in fitting and updating the kriging model. Formally, the EI step in Algorithm \ref{algorithm} is replaced with using $(x^{k+1}, \ell^{k+1})= (x^{k+2}, \ell^{k+2}) = \ldots =(x^{k+r}, \ell^{k+r}) $  and re-computing the EI score once every $r$ ground-level iterations.

\begin{figure}[ht]
  \centering
  \includegraphics[height=2.7in,trim=0.1in 0.1in 0.1in 0.1in]{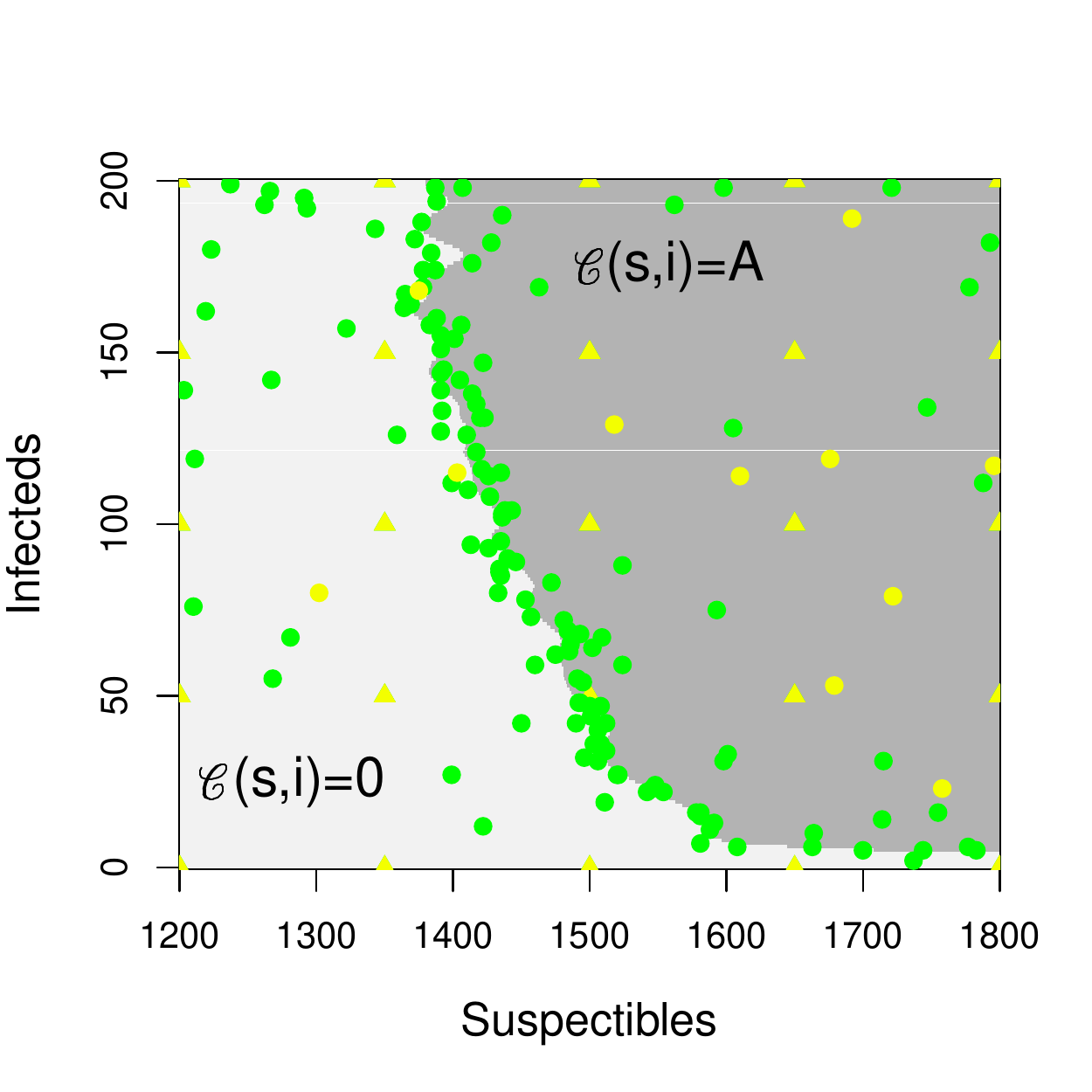} 
  \caption{Fitted response boundary  $\partial \CC$ for the epidemic response example using the Gap-SUR expected improvement metric. The scatterplot indicates the design $\ZZ^{(K)}$ for $K=200$; triangles indicate the initial design $\ZZ^{(K_0)}$, and circles the adaptively placed $(s,i)^{K_0:K}$ (green: $Y_0$; yellow: $Y_A$). \label{fig:epi-map}}
\end{figure}

For our study  we set  $M=2000$, $\beta^0 = 0.75, \beta^A=0.5$, $\gamma=0.5$ with intervention cost of $C^I=0.25$ per susceptible. Figure \ref{fig:epi-map} shows the resulting decision boundary $\partial \CC$. In the dark region the relative cost of intervention is lower, and hence action is preferred. For example, starting at $I_0 = 10, S_0 = 1800$, without any action the outbreak would affect more than 40\% of the susceptible population (expected cost of about 800), while under social distancing the impact would be about 60  infecteds (leading to much lower total expected cost of $60+C^I S_0 \simeq 510$). In the light region, wait-and-see approach has lower expected costs. For example at $I_0=50, S_0=1400$, the expected number of new infecteds without any action is $385$ while the cost of countermeasures is bigger at $0.25 \times 1400 + 102 = 452$. Overall, Figure \ref{fig:epi-map} shows that the optimal decision is very sensitive to the current number of susceptibles $S_0$. This feature is due to the fact that outbreaks are created when the infection rate dominates the recovery (reproductive ratio $\mathcal{R}_0 := (\beta^0/\gamma) (S_0/ M)$ above 1). Hence, for a pool with more than 85\% susceptibles ($S_0 > 1700$), the initial growth rate satisfies $\beta^0 S_0/ M > \gamma$ and is likely to trigger an outbreak. However, as $S$ is lowered, the region where $\beta^0 S_0/M \simeq \gamma$ is approached, which makes social distancing unnecessary, as outbreak likelihood and severity diminishes. In particular, Figure \ref{fig:epi-map} shows than no action is undertaken for $S_0 < 1350$. In the intermediate region, there is a nontrivial classifier boundary for determining $\CC(s,i)$.

\begin{figure}[ht]
  \centering
  \begin{tabular}{cc}
  \includegraphics[width=0.44\textwidth]{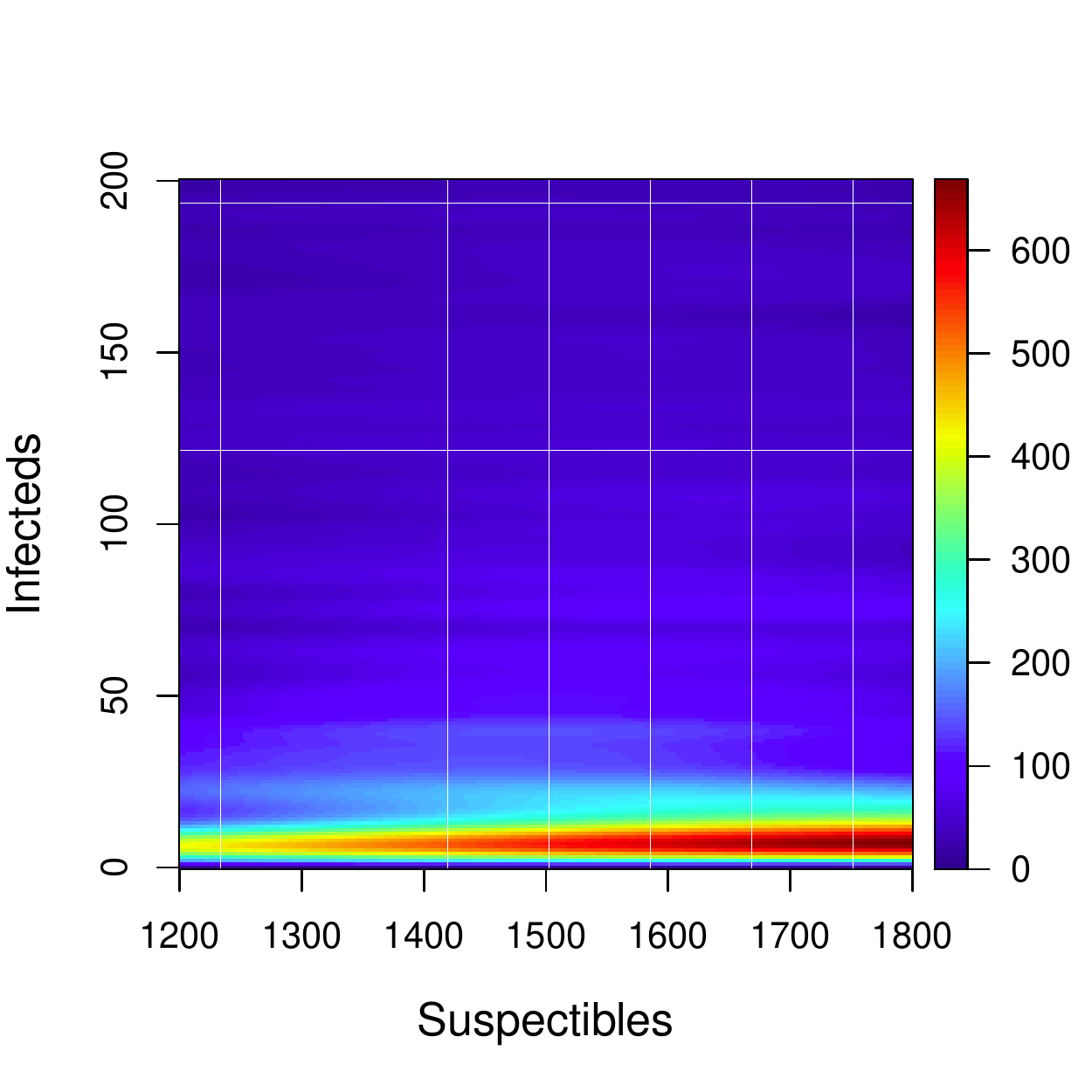} &
  \includegraphics[width=0.44\textwidth]{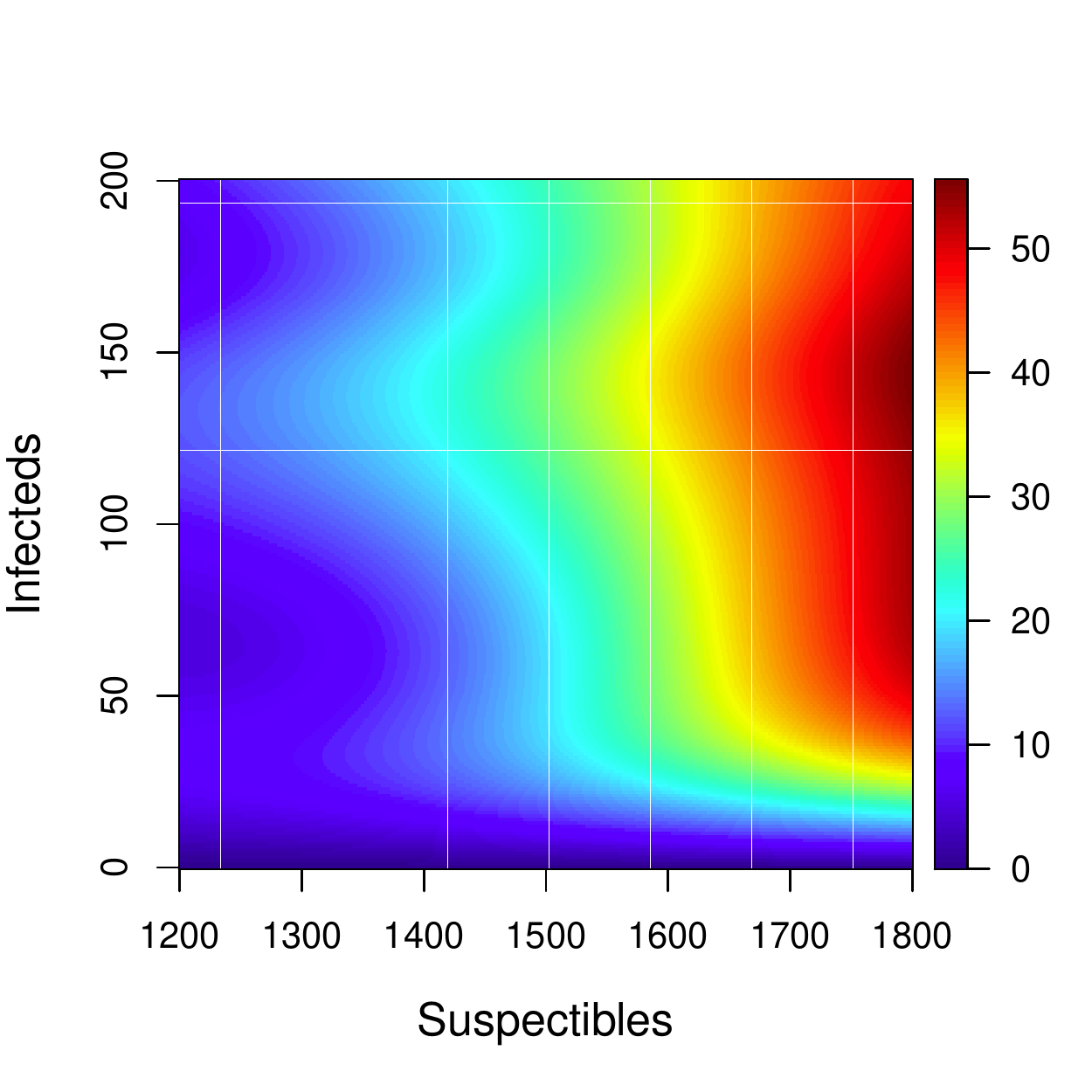}\\
  $\widetilde\sigma_{0}(s,i)$ & $\widetilde\sigma_{A}(s,i)$ \end{tabular}
  \caption{Estimated noise standard deviations $\widetilde{\sigma}_\ell(s,i)$ for the epidemic response example in the no-countermeasures (left panel, $\ell=0$) and action (right panel, $\ell=A$) regimes. Note the different color scales of the two panels, with $\sigma_0(\cdot) \gg \sigma_A(\cdot)$ for all $(s,i)$. \label{fig:varsurf}}
\end{figure}

Figure \ref{fig:epi-map} was generated by building an adaptive design using the Gap-SUR acquisition function and a total of $K=200$ design sites, with $r=100$ batched samples at each site. The input space was restricted to $\XX = \{ s \in \{1200, \ldots, 1800\}, i \in \{0, 200\} \}$.  The initial design $\ZZ^{(K_0)}$ included $50 = 25 \times 2$ sites on the same rectangular $5 \times 5$ lattice for each of $Y_0, Y_A$.
 In this example, the noise levels $\sigma^2_\ell(s,i)$ are highly state-dependent, see Figure \ref{fig:varsurf}. The $\mu_0$ surface has much higher noise, with largest $\sigma^2_0(s,i)$ for $(s,i) \simeq (1800,5))$, whereas $\mu_A$ has largest noise  in the top right corner. As a result, $\ZZ^{(K)}$ contains mostly samples from $Y_0$ and is denser towards the bottom of the Figure.

%
%

\section{Conclusion}\label{sec:conclude}
In this article we have constructed several efficient sequential design strategies for the problem of determining the minimum among $L \ge 2$ response surfaces. Our Gap-SUR heuristic connects \eqref{def_cal} to contour-finding and Bayesian optimization, providing a new application of the stepwise uncertainty reduction framework \cite{ChevalierPicheny13}. Our Gap-UCB heuristic mimics multi-armed bandits by treating all possible sampling pairs in $\mathcal{X}\times \mk{L}$ as arms, and trying to balance arm exploration and exploitation.

Our approach is based on the kriging framework, but this is  primarily for convenience and is not crucial. To this end, instead of a Bayesian formulation, one could use a maximum-likelihood method to fit $\widehat{\mu}_\ell(\cdot)$, replacing the posterior $\myM_\ell(x)$ with the point estimator and its standard error. Hence, many other regression frameworks could be selected. However, computational efficiency and the sequential framework place several efficiency restrictions on possible ways to modeling $\mu_\ell(\cdot)$. On the one hand, we need strong consistency, i.e.~the convergence of the respective classifier $\hat{\CC}^{(K)} \to \CC$ as $K\to \infty$. In particular, the regression method must be nonparametric and localized. On the other hand, we wish for a sequential procedure that allows for efficient updating rules in moving from $\hat\CC^{(k)}$ to $\hat\CC^{(k+1)}$. Lastly, in practical settings further challenges such as heteroscedasticity, non-Gaussian samplers $Y_\ell$, and heterogenous structure of the response surface are important.


One suitable alternative to GP's is local regression or Loess \cite{Loess}, which is a nonparametric  regression framework that fits pointwise linear regression models for $\mu_\ell(x)$. Loess is efficient and well-suited for heteroscedastic contexts with unknown noise distributions as in Section \ref{sec:epi}. It also automatically generates the posterior mean and variance of the fit (allowing to use the derived formulas based on $\widehat\mu_\ell(x)$ and $\delta_\ell(x)$). However, Loess is not updatable, creating computational bottlenecks if many design augmentation iterations are to be used. At the same time fitting is extremely fast, so depending on the implementation it might still be competitive with more sophisticated methods. In this spirit, piecewise linear regression (which first partitions $\XX$ into several cells and then carries out least-squares regression in each cell) is updatable via the Sherman-Morrison-Woodbury formulas and could be employed if there is a clear partitioning strategy available.

\red{We further note that GP kriging is just a convenient interim {surrogate} for building the experimental design. Consequently, once $\ZZ$ is generated, one could switch to a different response surface model to build a final estimate of the $\mu_\ell$'s and hence $\hat{\CC}$. For example, the treed GP approach \cite{tgpPackage} allows for a higher-fidelity fit for the response surfaces  when the underlying smoothness (specified by the covariance kernel) strongly varies across $\XX$. Because treed GP models are expensive to fit, one could compromise by using vanilla GP during DoE and treed GP for the final estimate of $\hat{\CC}$.}



Another fruitful extension would be to investigate ranking algorithms in the fixed confidence setting. As presented, the  sequential ranking algorithm is in the fixed budget setting, augmenting the design until a pre-specified size $K$. Practically, it is often desirable to prescribe adaptive, data-driven termination by targeting a pre-set confidence level. A good termination criterion should take both accuracy and efficiency into account, ensuring the accuracy of $\widehat{\mu}_\ell(x)$ and also anticipating low information gain from further sampling steps.  One proposed termination criterion is to keep track of the evolution of the empirical loss
$\mc{EL}( \hat{\CC}^{(k)})$,
and terminate once $\mc{EL}( \hat{\CC}^{(k)})-\mc{EL}( \hat{\CC}^{(k+1)})$ is small enough. This is equivalent to minimizing $L_k  :=  \mc{EL}( \hat{\CC}^{(k)}) + \underline{\eps} k$, where $\underline{\eps} > 0$ is a parameter for cost of simulations; the more we care about efficiency, the larger the $\underline{\eps}$ is.
When the design size $k$ is small, the first term will dominate, so $L_k$ is expected to first decrease in $k$. As $k \to \infty$, the rate of improvement in the loss function shrinks so that eventually $L_k$ will be increasing.
However, we find that $\mc{EL}( \hat{\CC}^{(k)})$ is quite noisy, especially if the kriging models are re-trained across stages. In that sense, the termination criterion needs to be robust enough to generate sufficiently strong (ad hoc) guarantees that a certain tolerance threshold has truly been achieved.

\printbibliography

\end{document}